\documentclass{article}

\usepackage{PRIMEarxiv}

\usepackage[utf8]{inputenc} 
\usepackage[T1]{fontenc}    
\usepackage[numbers,sort&compress]{natbib}
\usepackage{hyperref}       
\hypersetup{
  hidelinks,
  pdftitle={BeliefGraph-JEPA: Structured Latent World Models for Action-Conditioned Time Series},
  pdfauthor={Yue Li, Kangqi Ni, Zhen Tan, Tianlong Chen}
}
\usepackage{url}            
\usepackage{booktabs}       
\usepackage{amsfonts}       
\usepackage{amsmath,amssymb}
\usepackage{microtype}      
\usepackage{graphicx}       
\usepackage{array}
\usepackage{listings}

\makeatletter
\newcommand{\captionoftype}[1]{\def\@captype{#1}\caption}
\makeatother

\lstdefinestyle{prompt}{
  basicstyle=\scriptsize\ttfamily,
  breaklines=true,
  breakatwhitespace=true,
  columns=fullflexible,
  keepspaces=true,
  frame=single,
  aboveskip=0.5em,
  belowskip=0.5em,
  xleftmargin=2pt,
  xrightmargin=2pt,
  framexleftmargin=3pt,
  framexrightmargin=3pt,
  showstringspaces=false
}

\newcommand{\method}{\textsc{BeliefGraph-JEPA}}
\newcommand{\gr}{\textsc{GR}}
\newcommand{\ho}{\textsc{HO}}
\newcommand{\dc}{\textsc{DC}}
\newcommand{\sgraph}{\textsc{SG}}
\newcommand{\ie}{\textsc{IE}}
\newcommand{\rbel}{\textsc{RB}}
\newcommand{\fr}{\textsc{FR}}
\newcommand{\flatroll}{\textsc{flat}}
\newcommand{\flatrollgr}{\textsc{flat-GR}}
\newcommand{\urgr}{\textsc{UR-GR}}
\newcommand{\rbelgr}{\textsc{RB-GR}}
\newcommand{\frgr}{\textsc{FR-GR}}
\newcommand{\hb}{\textsc{HB}}
\newcommand{\frgrgo}{\textsc{FR-GR-GO}}
\newcommand{\frgru}{\textsc{FR-GR-U}}
\newcommand{\learnfc}{\textsc{LFC}}
\newcommand{\idmark}{$^{\mathrm{id}}$}
\newcommand{\dcid}{\dc{}\idmark{}}
\newcommand{\ieid}{\ie{}\idmark{}}
\newcommand{\rbelid}{\rbel{}\idmark{}}
\newcommand{\rbelgrid}{\rbelgr{}\idmark{}}
\newcommand{\frid}{\fr{}\idmark{}}
\newcommand{\frgrid}{\frgr{}\idmark{}}
\newcommand{\pmstd}[2]{$#1_{\scriptscriptstyle\pm#2}$}
\newcommand{\bestpm}[2]{$\mathbf{#1}_{\scriptscriptstyle\pm#2}$}
\newcommand{\secondpm}[2]{$\underline{#1}_{\scriptscriptstyle\pm#2}$}

\title{BeliefGraph-JEPA: Structured Latent World Models for Action-Conditioned Time Series}

\author{
\textbf{Yue Li}$^{1}$ \quad
\textbf{Kangqi Ni}$^{2}$ \quad
\textbf{Zhen Tan}$^{3}$ \quad
\textbf{Tianlong Chen}$^{2}$ \\[2pt]
\mdseries\normalsize
$^{1}$Carnegie Mellon University
\qquad
$^{2}$University of North Carolina at Chapel Hill \\[-1pt]
$^{3}$Stevens Institute of Technology
}

\begin{document}

\maketitle

\begin{abstract}
Action-conditioned time-series forecasting requires accounting for how
future actions and exogenous forcings influence multiple targets through
partially observed effects with different delays and persistence.
Direct conditioning leaves the evolution and target-specific influence
of these effects implicit in the predictor, while static relational
graphs specify connections without tracking evolving effects.
This motivates representing future-driver influence through structured
latent states that evolve over the forecast horizon and route
information to individual targets.
We introduce \method{}, a structured latent world model that factorizes
driver influence into typed latent-effect states.
These states are rolled forward under future drivers and routed through
a graph to target-specific nodes, forming the predictive base of a
joint-embedding predictive architecture.
A capacity-controlled residual supplements this base with direct
driver information.
On four multi-target clinical, agricultural, environmental, and
industrial systems, the framework outperforms a range of pretrained
and supervised known-future-covariate baselines.
Matched controls isolate latent dynamics, future rollout, graph routing,
and residual capacity; future rollout and graph-first residual routing
improve forecasting across all four systems.
\end{abstract}

\section{Introduction}
\label{sec:intro}

Action-conditioned time-series forecasting predicts how future actions
and exogenous forcings influence multiple targets.
In clinical, agricultural, environmental, and industrial systems, a
propofol infusion, a ventilation setpoint, a chemical dose, or a
heating-valve command is a \emph{driver}: an input that acts through
delayed, persistent, mechanism-specific hidden effects.
Actions are controllable inputs; exogenous forcings are inputs the
system cannot set.

Those effects are only partially observed and target-specific: blood
pressure and anesthetic depth respond to the same infusion with different
lags, while indoor temperature, humidity, CO$_2$, and light evolve together
under actuators and outdoor weather.

\begin{figure*}[t]
  \centering
  \includegraphics[width=0.8\textwidth]{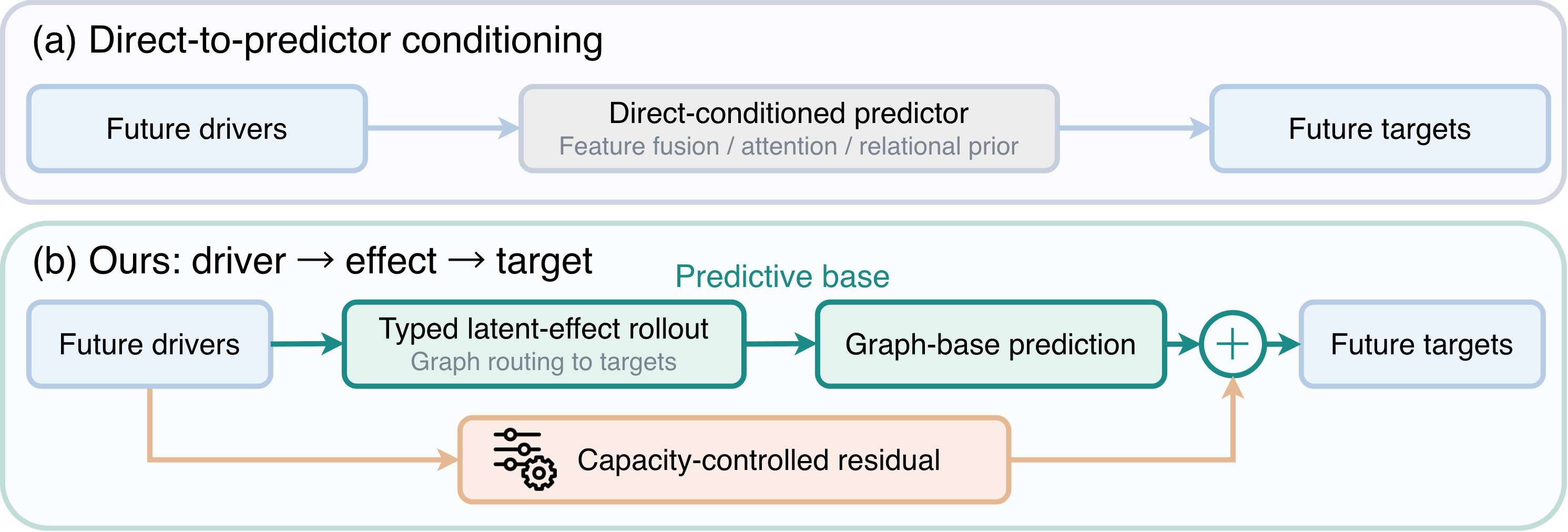}
  \caption{\textbf{Direct-to-predictor conditioning versus BeliefGraph-JEPA.}
  (a)~Drivers condition predictions without explicit typed latent-effect states.
  (b)~Drivers update typed latent-effect states, which are rolled to
  each horizon and routed to targets. Graph-base prediction of that
  trajectory is the predictive base; a capacity-controlled residual
  recovers omitted details.}
  \label{fig:overview}
\end{figure*}

The question is \emph{how to represent known future drivers whose effects
are delayed, mechanism-specific, and shared across multiple targets}.
Covariate-aware and pretrained
forecasters~\citep{salinas2020deepar,lim2021temporal,wang2024timexer,ansari2025chronos2}
and relational variants~\citep{qiu2025_dag} incorporate available drivers
as covariates, tokens, or relational inputs.
Other approaches combine action-conditioned prediction with rollout
or residual corrections~\citep{meng2026_cedar,liu2026_drift}.
Across these approaches, typed driver-mediated latent effects remain
implicit, as schematized in Figure~\ref{fig:overview}(a).
Graph-based forecasters model dependencies among observed
series~\citep{wu2019_graphwavenet,wu2020_mtgnn}, but adjacency alone
does not specify how driver-mediated latent effects evolve.
Latent world models already roll generic states forward under future
actions~\citep{watter2015_e2c,hafner2019_planet,hafner2025_dreamerv3}.
Our approach factorizes driver influence into typed latent-effect states,
rolls them forward under future drivers, and routes them through a graph
to target-specific nodes, using the structured trajectory as
the predictive base (Figure~\ref{fig:overview}(b)).

\method{} is a structured latent world model organized around a
pathway: driver $\rightarrow$ mechanism-specific latent effect
$\rightarrow$ observed target. A domain-informed typed graph supplies a relational prior.
Past drivers initialize mechanism-specific latent-effect states, and a
relation-aware GNN reads out their horizon-specific evolution as
target-specific graph representations.
Actions and exogenous forcings share one driver interface. The graph
represents the structured driver-mediated component of the partially
observed system, while the neural history representation captures
remaining unstructured factors. 
\emph{Graph-first residual routing} (\gr{}) uses these graph
representations together with observation history to form the
predictive base in JEPA space~\citep{assran2023self,ennadir2025_tsjepa}.
Direct driver tokens enter only through a gated residual adapter; a
residual-magnitude penalty and graph-base supervision constrain the
correction.

We evaluate this framework on four multi-target systems with shared
driver-mediated mechanisms: intraoperative vital
signs (VitalDB)~\citep{lee2022_vitaldb}, an autonomous
greenhouse (Greenhouse)~\citep{petropoulou2023_greenhouse}, wastewater nutrient
control (Wastewater)~\citep{mohammadi2024_wastewater}, and a district-heating
substation (PreDist)~\citep{roelofs2026_predist} (Table~\ref{tab:datasets}).

Our contributions are:
\begin{itemize}
  \item \textbf{Structured latent driver dynamics.} We factor
  future-driver influence into typed latent-effect states and roll
  those states forward before target-specific graph routing; that
  structured trajectory is the predictive base.
  On a matched-dynamics system, this rollout recovers the correct
  persistence ordering and exhibits pathway-specific impulse responses.
  \item \textbf{Graph-first residual routing.} The graph
  pathway is the predictive base. Direct drivers enter only through a
  gated residual, whose contribution is constrained by a
  residual-magnitude penalty and graph-base supervision.
  \item \textbf{Mechanism-oriented evaluation.} We use four
  multi-target systems with shared driver-mediated mechanisms and
  matched controls that separately test latent dynamics, future
  rollout, graph structure, and residual capacity.
\end{itemize}

\section{Related Work}
\label{sec:related}

\textbf{Action-conditioned forecasting and known-future covariates.}
Covariate-aware forecasters send known future actions or exogenous inputs
into the predictor. DeepAR, Temporal Fusion Transformer, and TiDE condition
on known future covariates~\citep{salinas2020deepar,lim2021temporal,das2023_tide};
TimeXer attends over historically observed exogenous
variables~\citep{wang2024timexer}. Foundation models such as Chronos and
TimesFM learn strong history extrapolation~\citep{ansari2024chronos,das2023decoder};
Chronos-2, ChronosX, and UniCA extend this line to future-known and
heterogeneous covariates~\citep{ansari2025chronos2,pineda2025_chronosx,han2026unica}.
ExoTimer integrates exogenous information with LLM-based
text representations~\citep{wu2026exotimer},
and DAG injects discovered endogenous--exogenous correlations into the same
forecast~\citep{qiu2025_dag}. 
CEDAR augments action-conditioned rollout with event
residuals~\citep{meng2026_cedar}; DRIFT adds recursive
action-conditioned corrections to direct forecasts~\citep{liu2026_drift}.
\method{} represents driver influence through typed latent
effects, rolled forward and routed to target-specific graph nodes,
with the structured trajectory as the predictive base.

\textbf{Latent world models, state-space models, and JEPA.}
Model-based control learns action-conditioned latent transitions
$z_{t+1}=F(z_t,a_t)$ and rolls them forward under future
actions~\citep{watter2015_e2c,hafner2019_planet,hafner2025_dreamerv3,zhou2024dino}.
Structured and selective state-space models provide efficient sequence backbones~\citep{gu2021_efficiently,gu2023_mamba}.
Joint-embedding predictive architectures predict in representation
space~\citep{assran2023self,bardes2024_vjepa,assran2025_vjepa2}, including
time series, command-conditioned control, and glucose representation
learning~\citep{ennadir2025_tsjepa,girgis2026_tsjepa,muhammad2026_cgmjepa}.
Phys-JEPA decomposes physical versus residual latent
content~\citep{nie2026_physjepa}; Delta-JEPA reconstructs actions from latent differences to encourage action-sensitive transitions~\citep{zhang2026_deltajepa}.
Our rollout is factorized by typed driver-mediated effect families
and routed to target-specific graph nodes, so future drivers enter as
inputs to structured latent mechanisms rather than only as predictor
covariates.

\textbf{Graph-structured and neuro-symbolic world models.}
Graph neural forecasters learn or assume adjacency among observed
variables~\citep{wu2019_graphwavenet,wu2020_mtgnn,cao2020_stemgnn};
GCGNet uses graph consistency over endogenous--exogenous correlations
~\citep{li2026_gcgnet}. LLM-constructed variable-level knowledge graphs add
a static semantic prior over the same series~\citep{sun2025_timemkg}.
Graph world models use message passing over structured states with action nodes~\citep{feng2025_gwm}; causal representation models impose
directed factor structure on static generative latents~\citep{yang2021_causalvae}.
Neuro-symbolic world models learn predicates and abstract dynamics for robot
planning~\citep{liang2025_visualpredicator,liang2026_exopredicator},
and medical trajectory world models simulate clinical courses in
language~\citep{mu2026_ehrworld}.
Our \method{} uses a typed graph to organize the
driver--effect--target pathway for action-conditioned forecasting.

\section{Problem Formulation}
\label{sec:problem}

At forecast anchor $t$, the model is given an observed context
$X_{t-L_X+1:t}=(x_{t-L_X+1},\ldots,x_t)$ of length $L_X$, with
$x_\tau\in\mathbb{R}^{d_x}$ at each time $\tau$. The task is the
continuation $Y_{t+1:t+H}=X_{t+1:t+H}$. The system is driven by
$D_\tau=\{A_\tau,\,E_\tau\}$,
where $A_\tau$ contains controllable actions and $E_\tau$ contains
uncontrollable exogenous forcings. The model is conditioned on future
actions over the prediction horizon; in our retrospective experiments,
these are the logged actions. Future exogenous inputs, when used, hold the last context
value (Table~\ref{tab:datasets}). Drivers act through
delayed, mechanism-specific hidden effects.
Those effects are represented as typed latent states and evolved under future drivers.
We treat the setting as a partially observed controlled dynamical
system.
The model represents the structured driver-mediated component by typed
latent-effect states $Z_t^{F}$, a deterministic summary of past driver
effects that is evolved under future drivers and routed to observed
targets.
Remaining unstructured factors are captured by the neural history
representation.

Joint-embedding prediction is performed in representation
space~\citep{assran2023self,ennadir2025_tsjepa}. An online context
encoder $E_\theta$ maps the observed history to $H_t^{X}$. An
exponential-moving-average (EMA) target encoder $E_{\bar\theta}$
encodes the future sequence,
$z_{t+1:t+H}^{*}=\operatorname{sg}\!\bigl(E_{\bar\theta}(X_{t+1:t+H})\bigr)$,
with $\operatorname{sg}$ the stop-gradient. A predictor produces
$\hat z_{t+1:t+H}$ from $H_t^{X}$, past drivers, and available future
drivers $D_{t:t+H-1}$. An observation decoder maps $\hat z_{t+h}$ to
$\hat y_{t+h}$. The latent and observation objectives are
\begin{equation}
\label{eq:jepa}
  \mathcal{L}_{\mathrm{JEPA}}
  =
  \frac{1}{H}\sum_{h=1}^{H}
  \|\hat z_{t+h}-z_{t+h}^{*}\|_{2}^{2},
  \qquad
  \mathcal{L}_{\mathrm{obs}}
  =
  \frac{1}{H}\sum_{h=1}^{H}
  \|\hat y_{t+h}-y_{t+h}\|_{2}^{2}.
\end{equation}
The base objective is
$\mathcal{L}=\mathcal{L}_{\mathrm{JEPA}}+\lambda_{\mathrm{obs}}\,\mathcal{L}_{\mathrm{obs}}$.
Section~\ref{sec:method} instantiates the predictor by rolling
$Z_t^{F}$ under subsequent drivers and using the resulting graph trajectory
as the predictive base.

\section{Method}
\label{sec:method}

\begin{figure*}[t]
  \centering
  \includegraphics[width=\textwidth]{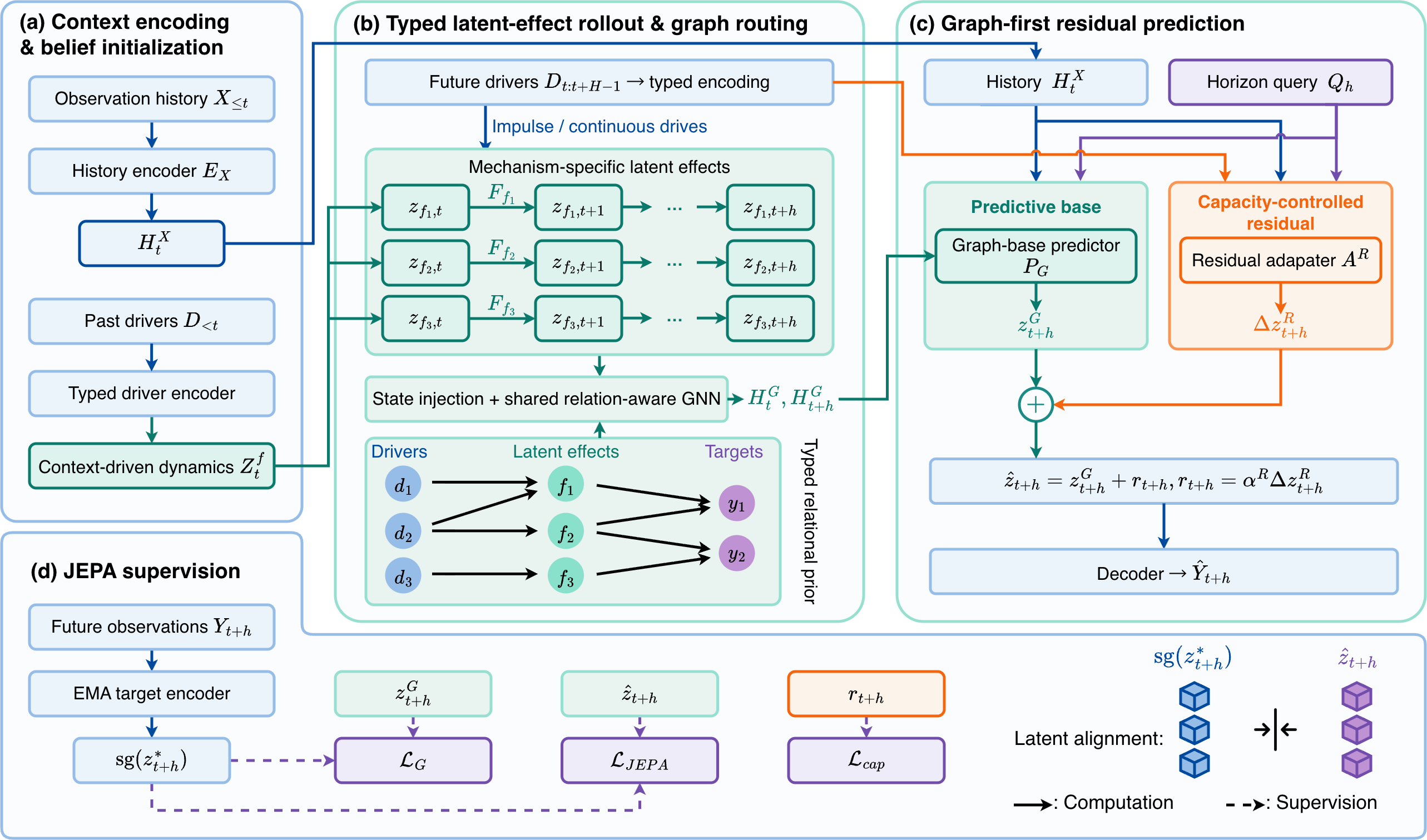}
  \caption{\textbf{BeliefGraph-JEPA: structured latent-effect rollout as the
  predictive base.}
  (a)~Past observations and past drivers initialize a history
  state $H_t^{X}$ and a family-effect state $Z_t^{f}$.
  (b)~Future drivers roll typed latent-effect states; a
  shared relation-aware GNN, guided by the typed
  driver--effect--target prior, produces the horizon-specific graph
  trajectory.
  (c)~$z_{t+h}^{G}$ is the predictive base; a
  capacity-controlled residual $\alpha^{R}\Delta z_{t+h}^{R}$ recovers
  omitted details.
  (d)~JEPA, graph-base, and capacity losses supervise the aligned
  latent.}
  \label{fig:method}
\end{figure*}

\method{} has two parts (Figure~\ref{fig:method}).
Section~\ref{sec:beliefgraph} turns a typed driver--effect--target prior
into typed latent-effect states and rolls them under future drivers.
Section~\ref{sec:gr} makes the resulting horizon-specific trajectory the
predictive base, with direct driver information entering as a
capacity-controlled residual.
Implementation details are in Appendix~\ref{app:arch}.

\subsection{Typed Latent-Effect Dynamics and Graph Rollout}
\label{sec:beliefgraph}

\textbf{Typed relational prior.}
We assume a sparse typed relational prior
$G_{\mathrm{static}}=(V_{D},V_{F},V_{Y},\mathcal{E})$
with $V_{D}=V_{A}\cup V_{E}$,
where $V_{A}$, $V_{E}$, $V_{F}$, and $V_{Y}$ are action, exogenous,
latent-effect, and target nodes.
$\mathcal{E}$ contains the directed
driver$\rightarrow$effect and effect$\rightarrow$target edges, each
with a relation label, sign, and confidence.
An instantaneous-effect head extends the readout to contemporaneous
algebraic targets on PreDist (Appendix~\ref{app:ie-head}).
The prior names which
driver families may update which latent effects, and which effects may
reach which targets, while sample-specific magnitude and temporal state
are learned. $V_{E}$ is a domain-selected exogenous subset; exogenous
channels not placed in $V_{E}$ remain available as direct predictor tokens.
Appendix~\ref{app:graph-prompt} records the frozen LLM-assisted
construction used in experiments.

\textbf{Unified driver representation.}
Actions and exogenous forcings share the interface $D_\tau=\{A_\tau,E_\tau\}$
and the same output dimension. Encoders keep three
attributes separate: numeric type (categorical versus continuous),
temporal modality (impulse versus continuous forcing), and role
(controllable versus exogenous). Each encoded driver yields a
token for the predictor and a latent drive $u_\tau$ that
updates mechanism-specific effect states. Impulse events are one-shot
jumps; continuous forcings remain active over physical time. Concurrent
drives of the same family and modality are summed. Exogenous graph nodes
are refreshed from observed history and, over the horizon, from the last
context value; persistence lives in the effect state.

\textbf{Current graph belief.}
For latent-effect family $f$ and a generic time $\tau$, impulse and
continuous drives update a recurrent state
\begin{equation}
\label{eq:effect-dyn}
  z_{f,\tau+1}
  =
  \lambda_f\odot z_{f,\tau}
  +
  \alpha_f^{\mathrm{imp}}\odot\phi_f^{\mathrm{imp}}(u_{f,\tau}^{\mathrm{imp}})
  +
  (1-\lambda_f)\odot\alpha_f^{\mathrm{cont}}\odot
  \phi_f^{\mathrm{cont}}(u_{f,\tau}^{\mathrm{cont}}),
\end{equation}
with $\sigma$ the sigmoid and $\lambda_f=\sigma(\rho_f)\in(0,1)$ a learned per-family
persistence, $\alpha_f^{\mathrm{imp}}$ and
$\alpha_f^{\mathrm{cont}}$ learned injection scales in $(0,1)$, and
$\phi_f^{\mathrm{imp}}$, $\phi_f^{\mathrm{cont}}$ pointwise maps of
the encoded drives. The impulse term is a state jump followed by
decay. The continuous term is stable first-order forcing;
$(1-\lambda_f)$ keeps a constant input from being amplified as
persistence approaches one. 
Past drivers $D_{t-L_A:t-1}$ are rolled through
Eq.~(\ref{eq:effect-dyn}) from a zero initial state to the forecast
anchor $t$. The packed
family-effect state $Z_t^{F}=(z_{f,t})_{f\in V_F}$ factorizes driver
influence by mechanism family.
The latent-effect window $L_A$ equals the context length $L_X$.

At the forecast anchor, a gate injects each family state
into its static node embedding $e_f^{\mathrm{static}}$,
\begin{equation}
\label{eq:inject}
  h_{f,\tau}^{(0)}
  =
  e_f^{\mathrm{static}}+\gamma_f W_f^{G}z_{f,\tau},
  \qquad
  \gamma_f=\sigma(\delta_f).
\end{equation}
The injection $\gamma_f W_f^{G}z_{f,\tau}$ keeps the magnitude of
$z_{f,\tau}$, which encodes accumulation under the recurrence.
History updates $z_{f,\tau}$ at every step and applies
Eq.~(\ref{eq:inject}) to the anchor state $z_{f,t}$.
$H_t^{(0)}$ initializes every node: effect nodes carry this injection,
and action and exogenous nodes carry their encoded driver features.
A two-layer relation-aware
GNN~\citep{scarselli2009_gnn,schlichtkrull2018_rgcn} then
propagates along $G_{\mathrm{static}}$,
$H_t^{G}=\operatorname{GNN}(G_{\mathrm{static}},H_t^{(0)})$.
Action and exogenous neighbors of an effect node send role-conditioned
messages; the two aggregates are concatenated before the effect update.
Target nodes receive messages from these effect representations.
Thus $H_t^{G}$ is
the GNN readout of $Z_t^{F}$, and the current graph belief is the pair
$(Z_t^{F},H_t^{G})$.

\textbf{Future latent-effect rollout.}
$Z_t^{F}$ summarizes drivers through time $t{-}1$. The same per-family
transition (Eq.~\ref{eq:effect-dyn}) is applied causally under subsequent
drivers. Writing $\mathcal{F}$ for that map on the packed family state,
$Z_{t+h}^{F}=\mathcal{F}(Z_{t+h-1}^{F},D_{t+h-1})$ for $h=1,\ldots,H$,
the horizon-$h$ state is obtained from $Z_t^{F}$ and $D_{t:t+h-1}$.
Each $Z_{t+h}^{F}$ is injected by Eq.~(\ref{eq:inject}), and action and
exogenous nodes at that step carry encodings of $D_{t+h-1}$. The same GNN
returns the horizon-specific readout $H_{t+h}^{G}$. Latent recurrence is
sequential and lightweight. The $H$ GNN evaluations are independent given
$\{Z_{t+h}^{F}\}$ and these driver encodings, and are batched over the
horizon. Future exogenous inputs hold the last context value
(Section~\ref{sec:problem}).

History $H_t^{X}$, the anchor graph $H_t^{G}$, and the horizon readout
$H_{t+h}^{G}$ enter a predictor $P$ with encoded tokens of the future
drivers $D_{t:t+h-1}$,
$\hat z_{t+h}=P(H_t^{X},\,H_t^{G},\,H_{t+h}^{G},\,D_{t:t+h-1},\,Q_h)$,
where $Q_h$ is a learned query embedding for horizon $h$.
The horizon queries are decoded jointly, with query $h$ using
$H_{t+h}^{G}$ and the driver prefix $D_{t:t+h-1}$.
We call this architecture \emph{future rollout} (\fr{}).

\subsection{Structured Rollout as the Predictive Base}
\label{sec:gr}

\fr{} lets the horizon graph readout and encoded future driver tokens
enter one predictor.
\gr{} designates the graph-mediated trajectory as the predictive base
$z_{t+h}^{G}=P_G(H_t^{X},\,H_t^{G},\,H_{t+h}^{G},\,Q_h)$.
The base $z_{t+h}^{G}$ is a JEPA-space prediction. History $H_t^{X}$ and
the anchor graph $H_t^{G}$ enter $P_G$ with the horizon readout
$H_{t+h}^{G}$. Future drivers reach this base through the rollout
$Z_{t+h}^{F}=\mathcal{F}(Z_{t+h-1}^{F},D_{t+h-1})$ and the subsequent GNN, so
$P_G$ conditions on them through $H_{t+h}^{G}$. Encoded tokens of
$D_{t:t+h-1}$ enter a residual adapter $A_R$ with a
stop-gradient copy of this base,
$\Delta z_{t+h}^{R}=A_R(H_t^{X},\,D_{t:t+h-1},\,\operatorname{sg}(z_{t+h}^{G}),\,Q_h)$.
The prediction is then
\begin{equation}
\label{eq:gr}
  \hat z_{t+h}
  =
  z_{t+h}^{G}
  +
  \alpha^{R}\Delta z_{t+h}^{R},
  \qquad
  \alpha^{R}
  =
  \alpha^{R}_{\max}\,\sigma(a),
\end{equation}
where $\alpha^{R}_{\max}<1$ is a fixed cap and $a$ is a learned scalar.
The residual adapter is zero-initialized, so
training begins from $\hat z_{t+h}\approx z_{t+h}^{G}$.

Graph-base supervision trains the base to match the JEPA target on its
own, and a residual-magnitude penalty keeps the residual a correction,
\begin{equation}
\label{eq:lg}
  \mathcal{L}_{G}
  =
  \frac{1}{H}\sum_{h=1}^{H}
  \|z_{t+h}^{G}-z_{t+h}^{*}\|_{2}^{2},
  \qquad
  \mathcal{L}_{\mathrm{cap}}
  =
  \frac{1}{H}\sum_{h=1}^{H}
  \|\alpha^{R}\Delta z_{t+h}^{R}\|_{2}^{2}.
\end{equation}
The training objective extends the base objective to
$\mathcal{L}=\mathcal{L}_{\mathrm{JEPA}}+\lambda_{G}\,\mathcal{L}_{G}
+\lambda_{\mathrm{cap}}\,\mathcal{L}_{\mathrm{cap}}
+\lambda_{\mathrm{obs}}\,\mathcal{L}_{\mathrm{obs}}$.
Residual reliance is the magnitude of the gated correction relative to
the combined magnitude of the base and correction,
\begin{equation}
\label{eq:rho}
  \rho_{R}
  =
  \frac{1}{H}\sum_{h=1}^{H}
  \frac{\|\alpha^{R}\Delta z_{t+h}^{R}\|_{2}}
  {\|z_{t+h}^{G}\|_{2}+\|\alpha^{R}\Delta z_{t+h}^{R}\|_{2}+\varepsilon}.
\end{equation}
where $\varepsilon>0$ is a small constant. Because $A_R$ learns the scale
of $\Delta z_{t+h}^{R}$, reliance is read from $\rho_{R}$ rather than from
the gate $\alpha^{R}$. The main method is
\frgr{}, which uses future rollout as the graph base and a
capacity-controlled residual on that base.

\section{Experiments}
\label{sec:experiments}

Table~\ref{tab:datasets} lists the four multi-target protocols, in each
of which several targets share typed driver-mediated mechanisms. For
retrospective evaluation, all action-conditioned methods receive the
same logged future actions. Future exogenous inputs, when present,
hold the last context value over the horizon. Sampling grids and forecast windows are in
Table~\ref{tab:app-multi-windows}; preprocessing and training settings
are in Appendix~\ref{app:preprocess} and Appendix~\ref{app:train}.

\begin{table}[t]
  \centering
  \caption{Multi-target forecasting protocols.
  $^{\dagger}$PreDist uses an instantaneous-effect head for algebraic
  heat power and flow (Appendix~\ref{app:ie-head}).}
  \label{tab:datasets}
  \scriptsize
  \setlength{\tabcolsep}{2.6pt}
  \resizebox{\columnwidth}{!}{%
  \begin{tabular}{@{}llll@{}}
    \toprule
    Dataset & Targets & Actions & Exogenous \\
    \midrule
    VitalDB
      & MBP, HR, SpO$_2$, BIS (4)
      & propofol, remifentanil rates
      & --- \\
    Greenhouse
      & indoor T, RH, CO$_2$, PAR (4)
      & heating, vents, screens, light, CO$_2$ dosing
      & outdoor weather (8) \\
    Wastewater
      & DO, NH$_4$, PO$_4$ (3)
      & metal-salt dosing; process phases
      & inflow, water temperature \\
    PreDist$^{\dagger}$
      & supply/return T, heat power, flow (5)
      & valve, setpoint; pump, mode, 3-way valve
      & outdoor T, grid supply T \\
    \bottomrule
  \end{tabular}}
\end{table}

\textbf{Metrics and controls.}
Primary metrics are pooled normalized MAE and RMSE on the $[-1,1]$
target scale, reported as mean $\pm$ sample standard deviation over
three seeds $\{42,43,44\}$. Baselines are last-value persistence,
history-only Chronos-2 and TimesFM~2.5, and covariate-aware
Chronos-2-Cov, DAG, and TFT, the last three with oracle future actions and
future exogenous inputs held at the last context
value~\citep{ansari2025chronos2,das2023decoder,qiu2025_dag,lim2021temporal}.
The sections below isolate latent dynamics, future rollout, graph
structure, and residual capacity in turn.

\subsection{Overall forecasting performance}
\label{sec:results-main}

Table~\ref{tab:main} reports the four-system comparison; per-target
physical-unit errors are in Appendix~\ref{app:per-target}.
TFT is one of the most competitive external baselines in our experiments
and uses the same constructed-exogenous
and oracle-action future protocol as DAG and \method{}.
\frgr{} attains lower nMAE and nRMSE on Greenhouse, VitalDB, and PreDist;
TFT attains slightly lower nMAE on Wastewater, with nRMSE essentially tied.
Per-target gains vary across variables (Appendix~\ref{app:per-target}):
\frgr{} attains the lowest mean MAE and RMSE on Greenhouse indoor
temperature, humidity, and CO$_2$, and on VitalDB BIS;
on PreDist, it also improves over TFT on heat power and flow.

To isolate where those gains arise, a cumulative ladder adds one
module at a time up to \fr{}: history only (\ho{}), future-driver
tokens (\dc{}), a static typed GNN (\sgraph{}), instant effect nodes
without temporal persistence (\ie{}), and a current recurrent belief
without future rollout (\rbel{}).
Direct future-driver conditioning already provides a gain over
history-only prediction: \dc{} improves over \ho{} on all four
systems, most clearly on Greenhouse and Wastewater.
A static typed GNN adds little beyond those tokens.
Persistent latent-effect dynamics then improve consistently:
\rbel{} improves over \ie{} on every dataset, and rolling those
states forward under future drivers improves again:
\fr{} improves over \rbel{}.

\flatroll{} keeps the \fr{} predictor but replaces GNN message passing
with an unstructured aggregation of effect states.
\flatrollgr{} applies the same base--residual decomposition,
auxiliary base supervision, and residual-capacity penalty to that
unstructured readout.
\urgr{} keeps the \flatrollgr{} readout and residual routing, and
replaces the family-specific latent states with one shared state.
\rbelgr{} applies the same design to the current-belief model.
\flatroll{} stays close to \fr{}, indicating that much of the
forecasting benefit is retained without GNN message passing.
This design improves both readouts: \flatrollgr{} over \flatroll{},
and \frgr{} over \fr{}.
\frgr{} also improves over \rbelgr{}, supporting future rollout
under matched residual routing.
\urgr{} performs similarly to \flatrollgr{} overall.
\frgr{} achieves lower mean errors than \urgr{} on all four datasets; 
its gains over \flatrollgr{} are clearest on Greenhouse, 
smaller on PreDist, and limited on Wastewater and VitalDB.
These results indicate dataset-dependent gains from explicit graph
routing.
The remaining diagnostics use \frgr{} as the main method.

\begin{table*}[t]
  \centering
  \caption{Test nMAE / nRMSE ($\downarrow$) on the four multi-target
  systems, mean $\pm$ sample std over three seeds. Best in each column in
  bold, second-best underlined. External baselines, a cumulative ladder
  up to \fr{}, then unstructured-readout and \gr{} controls.
  $^{\dagger}$Uses the instantaneous-effect head
  (Appendix~\ref{app:ie-head}).}
  \label{tab:main}
  \small
  \setlength{\tabcolsep}{3.6pt}
  \resizebox{\textwidth}{!}{%
  \begin{tabular}{@{}lrrrrrrrr@{}}
    \toprule
    & \multicolumn{2}{c}{Greenhouse} & \multicolumn{2}{c}{Wastewater}
      & \multicolumn{2}{c}{VitalDB} & \multicolumn{2}{c}{PreDist$^{\dagger}$} \\
    \cmidrule(lr){2-3} \cmidrule(lr){4-5}
    \cmidrule(lr){6-7} \cmidrule(lr){8-9}
    Method & nMAE & nRMSE & nMAE & nRMSE & nMAE & nRMSE & nMAE & nRMSE \\
    \midrule
    Persistence
      & $0.1552$ & $0.2603$ & $0.2911$ & $0.4548$
      & $0.0655$ & $0.1285$ & $0.0677$ & $0.1115$ \\
    Chronos-2
      & $0.1572$ & $0.2620$ & $0.2349$ & $0.3836$
      & $0.0640$ & $0.1244$ & $0.0598$ & $0.1004$ \\
    Chronos-2-Cov
      & $0.1464$ & $0.2521$ & $0.1765$ & $0.3118$
      & $0.0639$ & $0.1253$ & $0.0554$ & $0.0986$ \\
    TimesFM 2.5
      & $0.1562$ & $0.2579$ & $0.2231$ & $0.3584$
      & $0.0611$ & $0.1177$ & $0.0563$ & $0.0951$ \\
    DAG
      & \pmstd{0.1552}{0.0011} & \pmstd{0.2441}{0.0038}
      & \pmstd{0.1761}{0.0022} & \pmstd{0.2957}{0.0006}
      & \pmstd{0.0629}{0.0001} & \pmstd{0.1173}{0.0001}
      & \pmstd{0.0512}{0.0004} & \pmstd{0.0835}{0.0011} \\
    TFT
      & \secondpm{0.1327}{0.0014} & \pmstd{0.2007}{0.0026}
      & \bestpm{0.1201}{0.0008} & \bestpm{0.2139}{0.0031}
      & \pmstd{0.0588}{0.0017} & \pmstd{0.1083}{0.0018}
      & \pmstd{0.0576}{0.0048} & \pmstd{0.0890}{0.0073} \\
    \midrule
    \ho{}
      & \pmstd{0.1599}{0.0015} & \pmstd{0.2324}{0.0014}
      & \pmstd{0.2212}{0.0028} & \pmstd{0.3293}{0.0017}
      & \pmstd{0.0607}{0.0002} & \pmstd{0.1097}{0.0000}
      & \pmstd{0.0622}{0.0015} & \pmstd{0.0958}{0.0018} \\
    \dc{}
      & \pmstd{0.1486}{0.0008} & \pmstd{0.2137}{0.0004}
      & \pmstd{0.1520}{0.0033} & \pmstd{0.2493}{0.0043}
      & \pmstd{0.0596}{0.0005} & \pmstd{0.1056}{0.0001}
      & \pmstd{0.0578}{0.0015} & \pmstd{0.0889}{0.0011} \\
    \sgraph{}
      & \pmstd{0.1461}{0.0005} & \pmstd{0.2121}{0.0012}
      & \pmstd{0.1461}{0.0034} & \pmstd{0.2420}{0.0036}
      & \pmstd{0.0604}{0.0000} & \pmstd{0.1059}{0.0002}
      & \pmstd{0.0570}{0.0023} & \pmstd{0.0883}{0.0028} \\
    \ie{}
      & \pmstd{0.1479}{0.0036} & \pmstd{0.2139}{0.0040}
      & \pmstd{0.1479}{0.0030} & \pmstd{0.2444}{0.0035}
      & \pmstd{0.0600}{0.0003} & \pmstd{0.1057}{0.0002}
      & \pmstd{0.0578}{0.0017} & \pmstd{0.0891}{0.0019} \\
    \rbel{}
      & \pmstd{0.1461}{0.0013} & \pmstd{0.2122}{0.0017}
      & \pmstd{0.1443}{0.0009} & \pmstd{0.2394}{0.0011}
      & \pmstd{0.0597}{0.0003} & \pmstd{0.1049}{0.0004}
      & \pmstd{0.0551}{0.0017} & \pmstd{0.0862}{0.0018} \\
    \fr{}
      & \pmstd{0.1392}{0.0028} & \pmstd{0.2030}{0.0020}
      & \pmstd{0.1304}{0.0018} & \pmstd{0.2230}{0.0025}
      & \pmstd{0.0587}{0.0002} & \pmstd{0.1040}{0.0002}
      & \pmstd{0.0536}{0.0021} & \pmstd{0.0816}{0.0029} \\
    \midrule
    \flatroll{}
      & \pmstd{0.1401}{0.0016} & \pmstd{0.2052}{0.0033}
      & \pmstd{0.1304}{0.0038} & \pmstd{0.2208}{0.0039}
      & \pmstd{0.0589}{0.0001} & \pmstd{0.1035}{0.0000}
      & \pmstd{0.0561}{0.0025} & \pmstd{0.0841}{0.0023} \\
    \flatrollgr{}
      & \pmstd{0.1344}{0.0008} & \pmstd{0.2004}{0.0022}
      & \pmstd{0.1250}{0.0023} & \secondpm{0.2157}{0.0042}
      & \bestpm{0.0580}{0.0002} & \secondpm{0.1034}{0.0002}
      & \pmstd{0.0511}{0.0020} & \secondpm{0.0776}{0.0013} \\
    \urgr{}
      & \pmstd{0.1331}{0.0014} & \secondpm{0.1989}{0.0022}
      & \pmstd{0.1293}{0.0018} & \pmstd{0.2208}{0.0026}
      & \secondpm{0.0583}{0.0003} & \pmstd{0.1035}{0.0001}
      & \secondpm{0.0506}{0.0002} & \secondpm{0.0776}{0.0004} \\
    \rbelgr{}
      & \pmstd{0.1403}{0.0019} & \pmstd{0.2086}{0.0017}
      & \pmstd{0.1326}{0.0009} & \pmstd{0.2249}{0.0016}
      & \secondpm{0.0583}{0.0002} & \pmstd{0.1037}{0.0002}
      & \pmstd{0.0535}{0.0018} & \pmstd{0.0832}{0.0023} \\
    \frgr{}
      & \bestpm{0.1285}{0.0016} & \bestpm{0.1928}{0.0033}
      & \secondpm{0.1245}{0.0018} & \secondpm{0.2157}{0.0028}
      & \bestpm{0.0580}{0.0003} & \bestpm{0.1032}{0.0001}
      & \bestpm{0.0495}{0.0003} & \bestpm{0.0768}{0.0004} \\
    \bottomrule
  \end{tabular}}
\end{table*}

\subsection{Making structured rollout the predictive base}
\label{sec:results-routing}

Table~\ref{tab:routing} compares fusion controls.
\fr{} fuses the graph trajectory and direct future drivers in one
predictor.
A hard graph bottleneck (\hb{}) uses the same predictor but withholds
graph-covered future-driver tokens from the direct path.
\frgrgo{} evaluates the \frgr{} graph base at $\alpha^{R}=0$;
\frgru{} uses $\alpha^{R}=1$ and no residual-capacity penalty.

On all four datasets, \hb{} matches or slightly improves upon \fr{}:
graph-covered future driver information can be carried through the
rolled graph pathway without those tokens on the direct path.
The graph-only readout \frgrgo{} stays close to \fr{} on Greenhouse
and PreDist, so the rolled graph trajectory is already a
structured predictive base.
Removing the residual from \frgr{} raises error on every dataset,
so the direct correction still captures information beyond that base.
An unrestricted residual (\frgru{}) improves over \fr{} on every
dataset: graph-first prediction helps even without a
residual-capacity penalty.
Capacity control further improves Greenhouse, Wastewater, and
PreDist, while \frgr{} and \frgru{} remain comparable on VitalDB.
For \frgr{}, the relative residual magnitude $\rho_{R}$ is
$1.4$--$3.9\%$ (Appendix~\ref{app:residual}).
\frgr{} also outperforms \hb{} on all four datasets: a
capacity-controlled residual complements the graph base better
than a hard bottleneck.

\begin{table*}[t]
  \centering
  \caption{Routing and bottleneck diagnostics.}
  \label{tab:routing}
  \small
  \setlength{\tabcolsep}{3.6pt}
  \resizebox{\textwidth}{!}{%
  \begin{tabular}{@{}lrrrrrrrr@{}}
    \toprule
    & \multicolumn{2}{c}{Greenhouse} & \multicolumn{2}{c}{Wastewater}
      & \multicolumn{2}{c}{VitalDB} & \multicolumn{2}{c}{PreDist$^{\dagger}$} \\
    \cmidrule(lr){2-3} \cmidrule(lr){4-5}
    \cmidrule(lr){6-7} \cmidrule(lr){8-9}
    Method & nMAE & nRMSE & nMAE & nRMSE & nMAE & nRMSE & nMAE & nRMSE \\
    \midrule
    \fr{}
      & \pmstd{0.1392}{0.0028} & \pmstd{0.2030}{0.0020}
      & \pmstd{0.1304}{0.0018} & \pmstd{0.2230}{0.0025}
      & \pmstd{0.0587}{0.0002} & \pmstd{0.1040}{0.0002}
      & \pmstd{0.0536}{0.0021} & \pmstd{0.0816}{0.0029} \\
    \hb{}
      & \pmstd{0.1370}{0.0014} & \pmstd{0.2009}{0.0011}
      & \secondpm{0.1277}{0.0011} & \pmstd{0.2205}{0.0028}
      & \pmstd{0.0588}{0.0002} & \pmstd{0.1038}{0.0002}
      & \pmstd{0.0518}{0.0016} & \pmstd{0.0798}{0.0024} \\
    \midrule
    \frgrgo{}
      & \pmstd{0.1381}{0.0009} & \pmstd{0.2014}{0.0035}
      & \pmstd{0.1354}{0.0014} & \pmstd{0.2262}{0.0018}
      & \pmstd{0.0593}{0.0003} & \pmstd{0.1044}{0.0002}
      & \pmstd{0.0534}{0.0007} & \pmstd{0.0813}{0.0011} \\
    \frgru{}
      & \secondpm{0.1319}{0.0009} & \secondpm{0.1944}{0.0004}
      & \pmstd{0.1280}{0.0004} & \secondpm{0.2172}{0.0010}
      & \bestpm{0.0578}{0.0001} & \secondpm{0.1033}{0.0002}
      & \secondpm{0.0507}{0.0025} & \secondpm{0.0777}{0.0037} \\
    \rbelgr{}
      & \pmstd{0.1403}{0.0019} & \pmstd{0.2086}{0.0017}
      & \pmstd{0.1326}{0.0009} & \pmstd{0.2249}{0.0016}
      & \pmstd{0.0583}{0.0002} & \pmstd{0.1037}{0.0002}
      & \pmstd{0.0535}{0.0018} & \pmstd{0.0832}{0.0023} \\
    \frgr{}
      & \bestpm{0.1285}{0.0016} & \bestpm{0.1928}{0.0033}
      & \bestpm{0.1245}{0.0018} & \bestpm{0.2157}{0.0028}
      & \secondpm{0.0580}{0.0003} & \bestpm{0.1032}{0.0001}
      & \bestpm{0.0495}{0.0003} & \bestpm{0.0768}{0.0004} \\
    \bottomrule
  \end{tabular}}
\end{table*}

\subsection{What does graph structure contribute?}
\label{sec:results-graph}

\begin{table}[t]
  \centering
  \caption{Graph-structure controls.}
  \label{tab:graph}
  \scriptsize
  \setlength{\tabcolsep}{2pt}
  \resizebox{0.72\linewidth}{!}{%
  \begin{tabular}{@{}lrrrr@{}}
    \toprule
    & \multicolumn{2}{c}{Greenhouse} & \multicolumn{2}{c}{Wastewater} \\
    \cmidrule(lr){2-3} \cmidrule(lr){4-5}
    Method & nMAE & nRMSE & nMAE & nRMSE \\
    \midrule
    \multicolumn{5}{@{}l}{\textit{A. Retraining}} \\
    Domain graph (\frgr{})
      & \bestpm{0.1285}{0.0016} & \bestpm{0.1928}{0.0033}
      & \bestpm{0.1245}{0.0018} & \bestpm{0.2157}{0.0028} \\
    Learned-FC (\learnfc{})
      & \pmstd{0.1357}{0.0013} & \pmstd{0.2029}{0.0010}
      & \pmstd{0.1256}{0.0007} & \secondpm{0.2170}{0.0011} \\
    Shuffled-slot
      & \secondpm{0.1309}{0.0043} & \secondpm{0.1935}{0.0055}
      & \secondpm{0.1254}{0.0007} & \pmstd{0.2189}{0.0014} \\
    \midrule
    \multicolumn{5}{@{}l}{\textit{B. Inference-time prior swap}} \\
    Domain $\to$ shuffled-slot
      & \pmstd{0.1356}{0.0014} & \pmstd{0.2013}{0.0005}
      & \pmstd{0.1610}{0.0019} & \pmstd{0.2639}{0.0014} \\
    Shuffled-slot $\to$ domain
      & \pmstd{0.1365}{0.0046} & \pmstd{0.2008}{0.0053}
      & \pmstd{0.1472}{0.0046} & \pmstd{0.2503}{0.0081} \\
    \bottomrule
  \end{tabular}}
\end{table}

Where Table~\ref{tab:main} varies whether readout is a typed GNN,
Table~\ref{tab:graph} holds the \frgr{} backbone fixed and varies
which typed graph is used, by changing it during retraining or only
at inference.
These controls use Greenhouse and Wastewater, where the domain
prior is sparse and a typed fully-connected alternative is
well-defined.
Panel~A retrains \frgr{} with a learned fully-connected graph
(\learnfc{}) and a pre-specified slot-shuffled prior that remaps
driver$\rightarrow$effect slot inputs while keeping
effect$\rightarrow$target edges fixed.
Panel~B swaps the prior only at evaluation, testing the
domain-graph checkpoint under the slot shuffle and the
slot-shuffled checkpoint under the domain graph.

The domain graph attains the lowest error on both datasets.
A learned fully-connected graph is worse, especially on Greenhouse,
supporting the sparse typed prior over learning connectivity from
a complete graph.
Slot-shuffled retraining stays close, indicating that an
alternative typed routing remains competitive after retraining.
In contrast, swapping the same assignment only at inference raises
error sharply, especially on Wastewater (nMAE $0.1245$ to $0.1610$);
the reverse swap also increases error.
The typed graph is therefore a relational inductive bias:
predictions use the routing learned during training, while an
alternative typed assignment remains competitive once retrained.
The remaining experiments examine how the routed belief evolves
under future drivers.

\subsection{Can future rollout recover matched latent-effect dynamics?}
\label{sec:results-synth}

\begin{figure}[t]
  \begin{minipage}[c]{0.50\linewidth}
    \centering
    \includegraphics[width=0.95\linewidth]{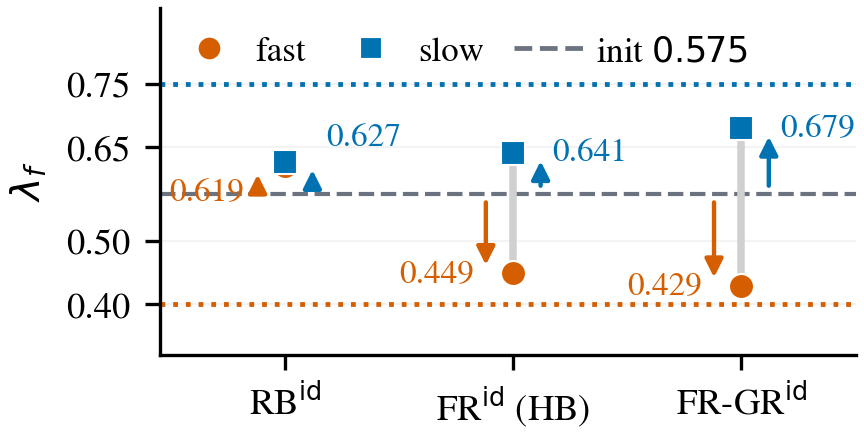}
    \caption{\textbf{Learned persistences under matched dynamics.}
    Learned $\lambda_f$ from init $0.575$ toward $(0.40,0.75)$ (seed $42$).}
    \label{fig:synth}
  \end{minipage}\hfill
  \begin{minipage}[c]{0.48\linewidth}
    \centering
    \captionoftype{table}{\textbf{Peak future-pulse responses under matched dynamics.}
    Peak $|I_{d\to y}|$, the maximum absolute response of
    target $y$ to a unit future pulse on driver $d$, on the two wired
    pathways and the off-path/null channels (seed $42$).}
    \label{tab:synth-impulse}
    \footnotesize
    \setlength{\tabcolsep}{2.8pt}
    \renewcommand{\arraystretch}{1.08}
    \begin{tabular}{@{}lccc@{}}
      \toprule
      & fast & slow & off-path \\
      \midrule
      Truth
        & $0.60$ & $0.25$ & $0$ \\
      \rbelid{}
        & $0$ & $0$ & $<0.1{\times}10^{-3}$ \\
      \frid{} (\hb{})
        & $0.37$ & $2{\times}10^{-4}$ & $9.6{\times}10^{-3}$ \\
      \frgrid{}
        & $0.43$ & $0.14$ & $8.2{\times}10^{-3}$ \\
      \bottomrule
    \end{tabular}
    \renewcommand{\arraystretch}{1}
  \end{minipage}
\end{figure}
Real forecasting data do not reveal ground-truth latent-effect states
or persistences, and Appendix~\ref{app:hyper} shows that multiple
persistence configurations can achieve nearly identical forecast
error.
We therefore probe capacity on a matched leaky-integrator with known
fast and slow latent mechanisms: a class-matched check of whether
future rollout recovers distinct persistences and pathway-specific
driver-to-target responses
(Figure~\ref{fig:synth}, Table~\ref{tab:synth-impulse}; Appendix~\ref{app:synthetic}).

The two wired effect families have
$\lambda_{\mathrm{fast}}=0.40$ and $\lambda_{\mathrm{slow}}=0.75$,
both initialized at $0.575$. Correct recovery therefore requires the
learned persistences to move in opposite directions. 
A third driver has no target path and serves as a null control. Superscript
\(\mathrm{id}\) denotes the matched protocol of
Appendix~\ref{app:synthetic}, which isolates future driver-to-latent
dynamics by restricting direct prediction shortcuts.

Figure~\ref{fig:synth} reports learned persistences against the
shared initialization and the two ground-truth scales.
Table~\ref{tab:synth-impulse} reports the corresponding peak
driver-to-target pulse responses.
Without future rollout, \rbelid{} leaves the two persistences nearly
equal and produces no pulse response.
\frid{} (\hb{}) recovers the correct ordering ($0.449<0.641$) and
halves MAE from $0.063$ to $0.031$, but only the fast pathway
responds ($0.371$); the slow pathway stays near zero.
\frgrid{} moves the estimates further toward the ground truth
($0.429<0.679$), reduces MAE to $0.013$, and recovers both wired
responses ($0.430$ and $0.141$), with off-path and null channels
near zero.
Among the compared variants, future rollout recovers the
correct persistence ordering, while the full \frgrid{} model
exhibits responses on both wired pathways, including the slow pathway.
Appendix~\ref{app:synthetic-mismatch} examines a saturating $\tanh$
mismatch: the slow-path response remains near zero at $40$ epochs
but increases at $80$ epochs, while both response magnitudes remain
underestimated.

\subsection{Do forecasts use future actions?}
\label{sec:results-perturb}

\begin{figure}[t]
  \begin{minipage}[t]{0.48\linewidth}
    \vspace{0pt}
    \centering
    \includegraphics[width=\linewidth]{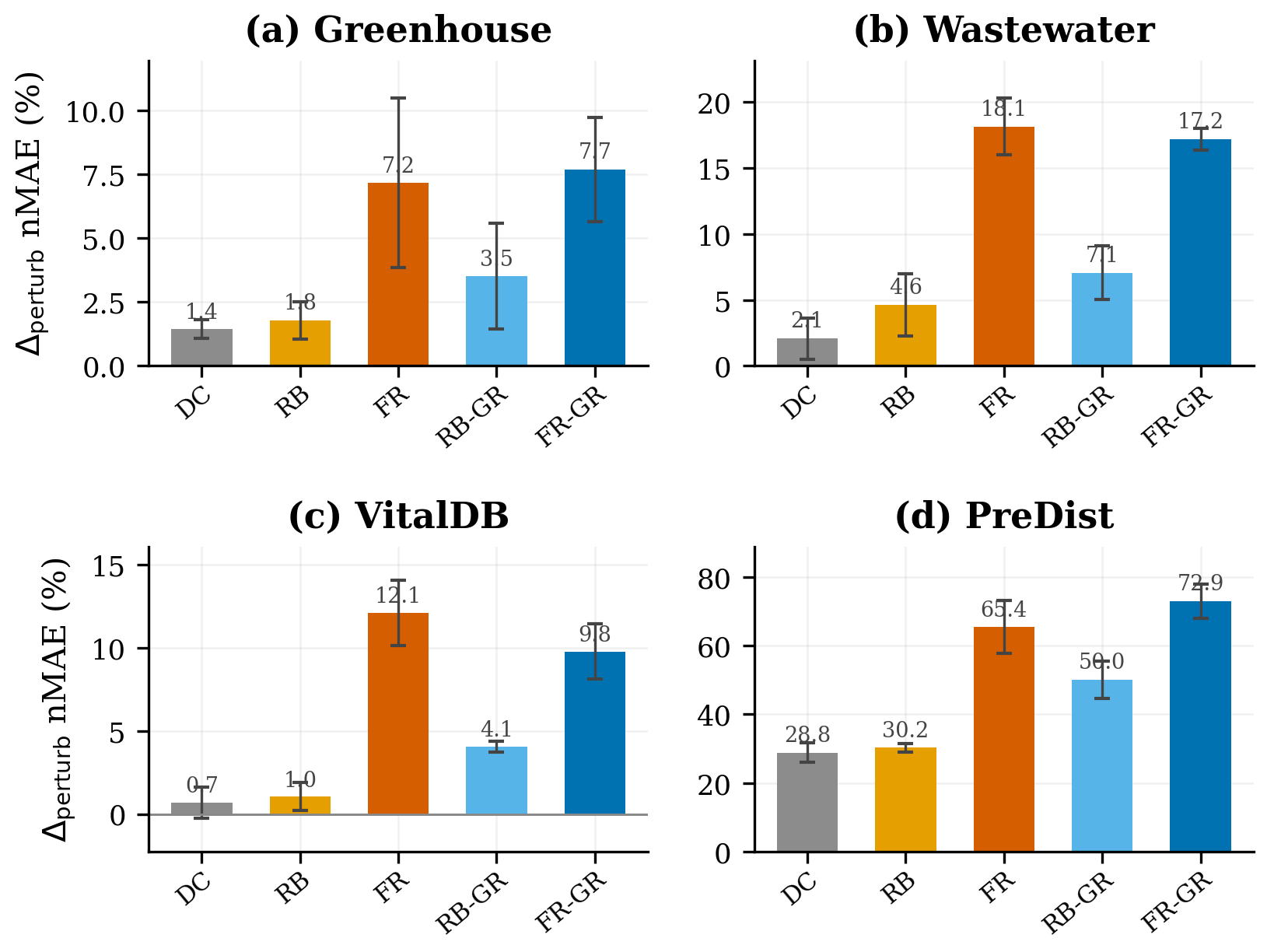}
    \caption{\textbf{Future-action perturbation.}
    $\Delta_{\mathrm{perturb}}$ after a
    second-half offset (last $90$\,min / $3$\,min / $3$\,h;
    Appendix~\ref{app:perturb}).
    Bars are mean $\pm$ sample std over three seeds.}
    \label{fig:perturb}
  \end{minipage}\hfill
  \begin{minipage}[t]{0.48\linewidth}
    \vspace{0pt}
    \centering
    \includegraphics[width=\linewidth]{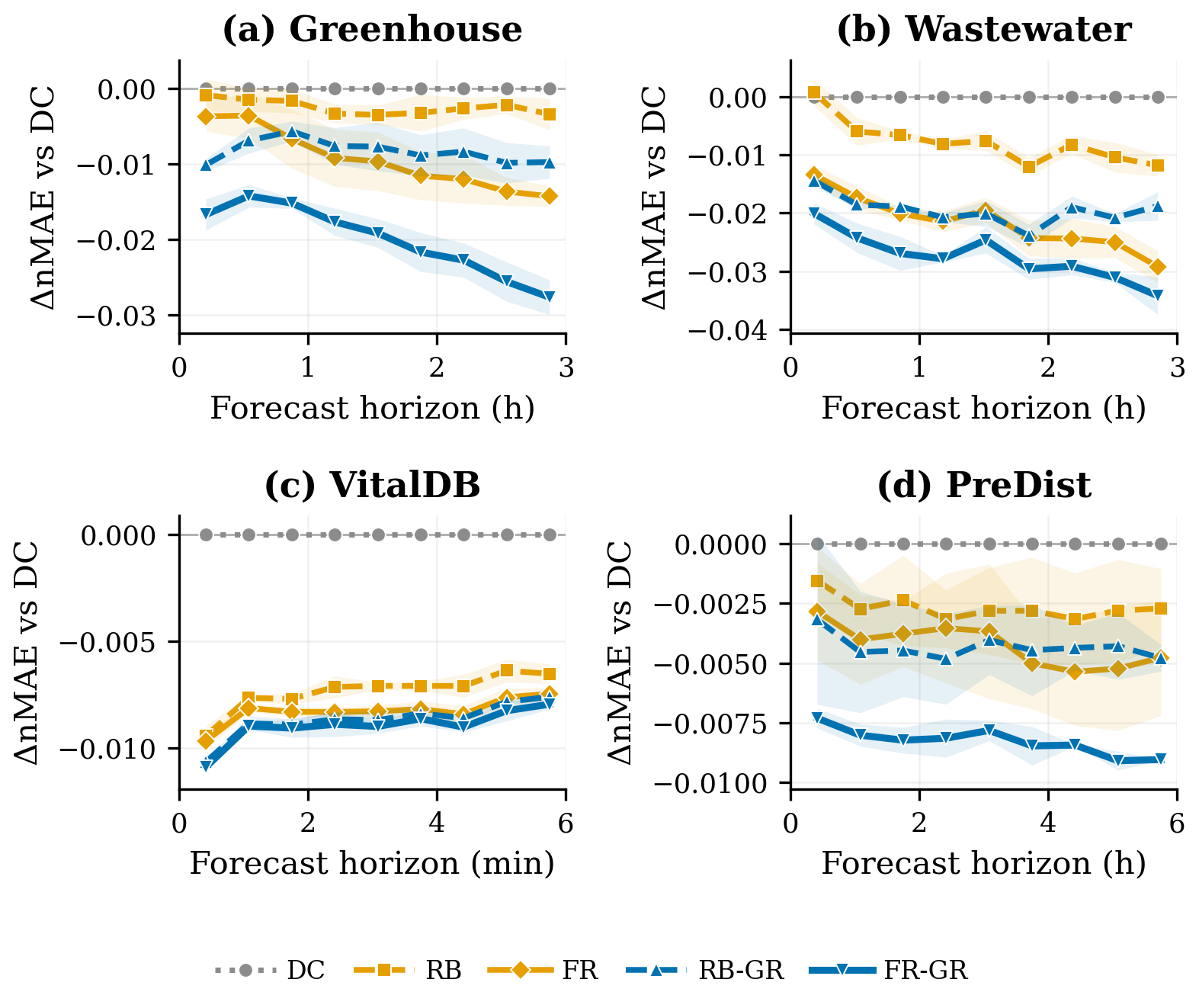}
    \caption{\textbf{Horizon-wise nMAE relative to \dc{}.}
    Mean $\pm$ sample std over three seeds; negative is lower
    error. The $x$-axis is physical lead time.}
    \label{fig:horizon}
  \end{minipage}
\end{figure}

Figure~\ref{fig:perturb} reports an input-dependence diagnostic:
error sensitivity to later actions
(Appendix~\ref{app:perturb}).
At evaluation we add a fixed signed offset to every
continuous future-action channel on the second half of the horizon,
hold exogenous inputs and categorical actions fixed, and report the percent nMAE increase
$\Delta_{\mathrm{perturb}}
=(\mathrm{nMAE}_{\mathrm{perturb}}-\mathrm{nMAE}_{\mathrm{clean}})
/\mathrm{nMAE}_{\mathrm{clean}}$
on the full window.

On Greenhouse, Wastewater, and VitalDB, \dc{} and \rbel{} show
only weak sensitivity to the second-half offset:
$\Delta_{\mathrm{perturb}}$ is $0.7$--$4.6\%$, despite \dc{}
receiving future-driver tokens directly. Explicitly rolling the
latent-effect belief under those later drivers produces a much
larger forecast-error increase: $7.2\%$, $18.1\%$, and $12.1\%$.
\rbelgr{} is intermediate ($3.5\%$, $7.1\%$, $4.1\%$): the residual
adapter sees the perturbed tokens, while the graph base remains the
current belief. \frgr{} matches or exceeds \fr{} on Greenhouse
($7.7\%$ versus $7.2\%$) and remains comparable on Wastewater
($17.2\%$ versus $18.1\%$), and remains an order of magnitude
above \rbel{} on VitalDB ($9.8\%$ versus $1.0\%$).
On PreDist, $\Delta_{\mathrm{perturb}}$ is already $28.8\%$ for
\dc{}. Future rollout still produces the largest error increase
($65.4\%$ for \fr{}, $72.9\%$ for \frgr{}).
Future rollout therefore makes predictions depend on later actions,
including offsets that token-only conditioning largely ignores.

\subsection{Horizon-wise benefit of future rollout}
\label{sec:results-horizon}

Figure~\ref{fig:horizon} reports per-horizon nMAE relative to \dc{}.
Negative values are lower error. \frgr{} is lowest at every lead time.
Across all four datasets, \fr{} and \frgr{} generally outperform
their current-belief counterparts, \rbel{} and \rbelgr{},
across lead times.
The gap widens with lead time on Greenhouse and Wastewater,
most steadily on Greenhouse.
On PreDist, \frgr{} remains lowest throughout.
On VitalDB's six-minute horizon, \frgr{} slightly improves over
\rbelgr{}.
Together, these comparisons support evolving latent effects under
future drivers, with the magnitude of the benefit varying across
datasets and lead times.

\section{Conclusion}
\label{sec:conclusion}

We presented \method{}, a structured latent world model for
action-conditioned forecasting.
Future drivers enter latent dynamics through typed latent-effect states
and graph-mediated rollout, with a capacity-controlled residual.
On four multi-target systems across distinct domains, the framework
outperforms a range of pretrained and supervised known-future-covariate
baselines.
Explicit graph routing provides additional, dataset-dependent gains.
Matched ablations isolate latent dynamics, future rollout, graph
routing, and residual capacity.
On a matched-dynamics system, the model recovers the correct persistence 
ordering and exhibits pathway-specific impulse responses.
On the real systems, later-action offsets change the forecast once
the latent-effect states are rolled forward.
Identifying latent mechanisms and extending the framework to action
planning remain directions for future work.

\bibliographystyle{unsrtnat}
\bibliography{references}

\appendix

\section{Datasets}
\label{app:datasets}

The main evaluation uses four multi-target controlled systems
(Table~\ref{tab:datasets}, Appendix~\ref{app:datasets-multi}), in
which several targets share typed driver-mediated mechanisms.
We apply \method{} to two
single-target glucose protocols
(Table~\ref{tab:app-single-datasets},
Appendix~\ref{app:datasets-single}), where the typed graph has one
observed sink and fewer effect-to-target routes.
Forecasting results for those protocols are in
Appendix~\ref{app:single-target}. Per-target breakdowns of the four
multi-target systems are in Appendix~\ref{app:per-target}. Train/val/test
splits and remaining preprocessing are in Appendix~\ref{app:preprocess}.

\subsection{Multi-target datasets}
\label{app:datasets-multi}

\textbf{VitalDB.}
Intraoperative vital signs from surgical cases~\citep{lee2022_vitaldb}.
Targets are mean arterial pressure, heart rate, pulse oxygen saturation, and
bispectral index. Continuous actions are propofol and remifentanil infusion
rates. There is no exogenous channel. The same infusion can affect pressure
and anesthetic depth with different lags, so the typed graph routes one
driver family to several targets.

\textbf{Greenhouse.}
The 3rd Autonomous Greenhouse Challenge~\citep{petropoulou2023_greenhouse}:
lettuce compartments under closed-loop climate control. Targets are indoor
air temperature, relative humidity, CO$_2$ concentration, and incoming PAR.
Continuous actions are heating-pipe temperature, leeward and windward vents,
thermal and shade screens, grow lights, and CO$_2$ dosing. Exogenous inputs
are outdoor weather. Future outdoor weather holds the last context value
over the horizon.

\textbf{Wastewater nutrient control.}
A biological nutrient-removal tank with online sensors and
dosing~\citep{mohammadi2024_wastewater}. Targets are dissolved oxygen,
ammonium, and phosphate. Continuous actions are metal-salt dosing channels;
categorical actions are inlet and outlet process phases. Exogenous inputs are
inflow rate and water temperature. Dosing and phase changes persist beyond a
single step and affect the three nutrient targets through shared process
mechanisms.

\textbf{PreDist.}
District-heating substations~\citep{roelofs2026_predist}. Targets include
secondary supply temperature, return temperatures, heat power, and flow.
Continuous actions are valve position and supply-temperature setpoint;
categorical actions include pump status, control mode, and a three-way
valve. Exogenous inputs are outdoor temperature and primary-grid supply
temperature. Heat power and flow are algebraic: their instantaneous values
are determined by contemporaneous temperatures and hydraulic drivers.
This protocol mixes delayed inertial targets with same-horizon algebraic
targets and uses an instantaneous-effect head
(Appendix~\ref{app:ie-head}).

Table~\ref{tab:app-multi-windows} lists the sampling grid and forecast
windows.

\begin{table}[ht]
  \centering
  \caption{Multi-target grids and forecast windows.}
  \label{tab:app-multi-windows}
  \footnotesize
  \setlength{\tabcolsep}{4pt}
  \begin{tabular}{lll}
    \toprule
    Dataset & Grid & Context, horizon \\
    \midrule
    VitalDB & 10\,s & 15\,min, 6\,min \\
    Greenhouse & 5\,min & 12\,h, 3\,h \\
    Wastewater & 2\,min & 6\,h, 3\,h \\
    PreDist & 10\,min & 12\,h, 6\,h \\
    \bottomrule
  \end{tabular}
\end{table}

\subsection{Single-target datasets}
\label{app:datasets-single}

Table~\ref{tab:app-single-datasets} summarizes the two single-target glucose
protocols. Both use the same driver--graph--rollout--residual framework as
the multi-target systems; there is no exogenous channel.

\begin{table}[t]
  \centering
  \caption{Single-target glucose protocols.}
  \label{tab:app-single-datasets}
  \footnotesize
  \setlength{\tabcolsep}{4pt}
  \begin{tabular}{llll}
    \toprule
    Dataset & Grid & Context, horizon & Actions \\
    \midrule
    CGMacros & 5\,min & 12\,h, 2\,h & sparse meal events \\
    Shanghai Diabetes & 15\,min & 12\,h, 2\,h
      & sparse meal, injection, and oral-drug events \\
    \bottomrule
  \end{tabular}
\end{table}

\textbf{CGMacros.}
Continuous glucose monitoring with meal logs~\citep{das2025_cgmacros}. The
sole target is interstitial glucose. Actions are sparse meal events with
macronutrient annotations; there is no exogenous channel and no dense
control input. The typed graph therefore has a single observed target and
fewer effect-to-target routes than the multi-target systems above.

\textbf{Shanghai Diabetes.}
Shanghai T1DM/T2DM CGM with meal, injection, and oral-drug
logs~\citep{zhao2023_shanghai}. The sole target is glucose. Actions are
sparse clinical events; there is no exogenous channel. As in CGMacros, the
graph topology is a single-target driver--effect--glucose routing structure.

\section{Dataset preprocessing}
\label{app:preprocess}

Windows never cross session boundaries. A session break is a contiguous
invalid-target gap (VitalDB, Greenhouse, Wastewater, PreDist) or a timestamp
gap of more than 30 minutes (CGMacros, Shanghai Diabetes). Short sessions and
windows whose target NaN ratio exceeds a dataset-specific threshold are
dropped. Target and exogenous channels are min-max normalized to $[-1,1]$
with fixed physical ranges. Continuous actions use channel-wise min-max.
Categorical actions remain integer indices. Nutrition and dose features on
the glucose datasets are z-scored from the training split only.

\textbf{VitalDB.}
Native recordings are on a 2\,s grid and are resampled to 10\,s for
training. Splits are by surgical case at 70/15/15 with seed 42: the same
case never appears in more than one split.

\textbf{Greenhouse.}
Splits are by team/compartment with seed 42: two teams are held out for
test, one for validation, and the remainder for training. Compartments from
the same team stay in the same split.

\textbf{Wastewater nutrient control.}
The source is a single plant, so splits are chronological by calendar month
at 70/15/15. Earlier months train, later months validate and test; there is
no random subject shuffle.

\textbf{PreDist.}
Splits are by substation at 70/15/15 with seed 42. Maintenance events are
not treated as actions.

\textbf{CGMacros.}
Native CGM is 1\,min and is resampled to 5\,min for training. Splits are by
subject at 70/15/15 with seed 42, stratified by A1c phenotype (healthy,
prediabetes, type 2 diabetes) so that each phenotype appears in every split
when the cohort is large enough.

\textbf{Shanghai Diabetes.}
Splits are by subject at 70/10/20 with seed 42, stratified by phenotype
(T1DM, T2DM) and then merged. Adjacent records more than 30 minutes apart
form distinct sessions.

\section{Single-target evaluation}
\label{app:single-target}

Table~\ref{tab:app-single} applies the \ho{}--\fr{} ladder and its
\gr{} variants from Table~\ref{tab:main} to the two glucose protocols of
Table~\ref{tab:app-single-datasets}. The typed graph has one observed
target, so there are fewer effect-to-target routes than on the four
multi-target systems. Metrics are physical-unit MAE and RMSE in
mg/dL. Foundation-model rows are single-run; remaining rows are
mean $\pm$ sample std over three seeds.

\begin{table}[t]
  \centering
  \caption{Test MAE / RMSE ($\downarrow$) in mg/dL on the two
  single-target glucose datasets. Foundation-model rows are single-run;
  remaining rows are mean $\pm$ sample std over three seeds. The
  \ho{}--\fr{} ladder and its \gr{} variants.}
  \label{tab:app-single}
  \footnotesize
  \setlength{\tabcolsep}{3.6pt}
  \begin{tabular}{@{}lrrrr@{}}
    \toprule
    & \multicolumn{2}{c}{CGMacros} & \multicolumn{2}{c}{Shanghai Diabetes} \\
    \cmidrule(lr){2-3} \cmidrule(lr){4-5}
    Method & MAE & RMSE & MAE & RMSE \\
    \midrule
    Persistence
      & $17.25$ & $28.11$ & $18.71$ & $29.55$ \\
    Chronos-2
      & $15.42$ & $25.38$ & $18.96$ & $30.56$ \\
    Chronos-2-Cov
      & $15.91$ & $26.31$ & $18.75$ & $30.37$ \\
    TimesFM 2.5
      & $14.89$ & $24.59$ & $17.25$ & $27.72$ \\
    DAG
      & \pmstd{13.37}{0.12} & \pmstd{21.52}{0.21}
      & \secondpm{15.88}{0.26} & \pmstd{25.05}{0.45} \\
    TFT
      & \secondpm{12.79}{0.21} & \pmstd{20.81}{0.39}
      & \bestpm{15.30}{0.02} & \bestpm{23.88}{0.13} \\
    \midrule
    \ho{}
      & \pmstd{14.72}{0.07} & \pmstd{23.42}{0.38}
      & \pmstd{17.08}{0.30} & \pmstd{25.80}{0.13} \\
    \dc{}
      & \pmstd{13.24}{0.15} & \pmstd{20.62}{0.15}
      & \pmstd{16.61}{0.14} & \pmstd{25.60}{0.25} \\
    \sgraph{}
      & \pmstd{13.20}{0.09} & \pmstd{20.69}{0.22}
      & \pmstd{16.69}{0.06} & \pmstd{25.49}{0.40} \\
    \ie{}
      & \pmstd{13.14}{0.04} & \pmstd{20.26}{0.13}
      & \pmstd{16.49}{0.05} & \pmstd{25.44}{0.12} \\
    \rbel{}
      & \pmstd{12.92}{0.20} & \pmstd{20.09}{0.27}
      & \pmstd{16.51}{0.25} & \pmstd{25.57}{0.34} \\
    \fr{}
      & \pmstd{12.88}{0.17} & \pmstd{20.18}{0.25}
      & \pmstd{16.33}{0.31} & \pmstd{25.27}{0.15} \\
    \midrule
    \rbelgr{}
      & \pmstd{12.88}{0.03} & \secondpm{19.99}{0.07}
      & \pmstd{16.24}{0.31} & \secondpm{25.01}{0.39} \\
    \frgr{}
      & \bestpm{12.76}{0.25} & \bestpm{19.95}{0.16}
      & \pmstd{16.52}{0.19} & \pmstd{25.41}{0.13} \\
    \bottomrule
  \end{tabular}
\end{table}

On CGMacros, future meal tokens produce the large step: \dc{} lowers
MAE from $14.72$ (\ho{}) to $13.24$. Persistent effects and future
rollout add smaller increments. \frgr{} attains the lowest MAE and
RMSE ($12.76$ and $19.95$), essentially tied with TFT ($12.79$ and
$20.81$) and ahead of DAG ($13.37$) and TimesFM~2.5
($14.89$). Relative to \ho{}, that is a $13\%$ MAE reduction.

On Shanghai Diabetes, actions are sparse typed clinical events
(meal, injection, oral drug) and the graph has a single glucose sink.
TFT attains the lowest MAE and RMSE ($15.30$ and $23.88$), ahead of
DAG ($15.88$). Within the shared ladder,
\rbelgr{} is lowest ($16.24$ MAE, $25.01$ RMSE) and \frgr{} is
$16.52$.

Table~\ref{tab:app-shanghai-residual} evaluates the graph base
$z^{G}$ ($\alpha^{R}{=}0$) and the full prediction
$\hat z=z^{G}+\alpha^{R}\Delta z^{R}$ on the same checkpoints.
Residual gain is $\mathrm{err}(z^{G})-\mathrm{err}(\hat z)$.
The residual of \rbelgr{} lowers MAE by $0.86$.
Future rollout improves the graph base by $0.32$
($z^{G}$ MAE $17.10$ to $16.78$); the residual of \frgr{} then
lowers MAE by only $0.25$.
The two mechanisms are substitutes for future-driver
information, and the current-belief residual is the stronger
route.
Future events occupy $40\%$ of Shanghai test windows, so
\rbelgr{} already reads those tokens, while rolled $Z^{F}$ is
empty on the complementary windows.

\begin{table}[t]
  \centering
  \caption{Shanghai residual gain. Graph-base error at
  $\alpha^{R}{=}0$ versus the full prediction
  $\hat z=z^{G}+\alpha^{R}\Delta z^{R}$ on the same
  \rbelgr{} and \frgr{} checkpoints as
  Table~\ref{tab:app-single}. Residual gain
  $\Delta=\mathrm{err}(z^{G})-\mathrm{err}(\hat z)$ is
  computed per seed. Mean $\pm$ sample std over three seeds.}
  \label{tab:app-shanghai-residual}
  \footnotesize
  \setlength{\tabcolsep}{3.2pt}
  \begin{tabular}{@{}lrrrrrr@{}}
    \toprule
    & \multicolumn{3}{c}{MAE} & \multicolumn{3}{c}{RMSE} \\
    \cmidrule(lr){2-4} \cmidrule(lr){5-7}
    Method
      & $z^{G}$ & $\hat z$ & $\Delta$
      & $z^{G}$ & $\hat z$ & $\Delta$ \\
    \midrule
    \rbelgr{}
      & \pmstd{17.10}{0.13} & \pmstd{16.24}{0.31} & \pmstd{0.86}{0.19}
      & \pmstd{25.81}{0.22} & \pmstd{25.01}{0.39} & \pmstd{0.81}{0.22} \\
    \frgr{}
      & \pmstd{16.78}{0.18} & \pmstd{16.52}{0.19} & \pmstd{0.25}{0.04}
      & \pmstd{25.62}{0.18} & \pmstd{25.41}{0.13} & \pmstd{0.21}{0.05} \\
    \bottomrule
  \end{tabular}
\end{table}

The same framework also applies to single-target action-conditioned forecasting.
With one observed sink the gain over \ho{} is smaller than in
Table~\ref{tab:main}, and the best \gr{} variant is not the same
on both protocols: \frgr{} on CGMacros, \rbelgr{} on Shanghai.
TFT is essentially tied with \frgr{} on CGMacros and is strongest on
Shanghai.
The strongest benefits of the full framework are observed in the
multi-target setting, where several targets share typed
driver-mediated mechanisms.
A plausible explanation is that multiple effect-to-target routes provide
the graph pathway with target-specific routing structure
that a single-sink graph does not provide.

\section{Per-target evaluation}
\label{app:per-target}

Table~\ref{tab:main} pools every target on the $[-1,1]$ scale.
Tables~\ref{tab:app-per-target-greenhouse}--\ref{tab:app-per-target-predist}
report physical-unit MAE and RMSE for each target, with the same methods
and seeds. Foundation-model rows are single-run; remaining rows are
mean $\pm$ sample std over seeds $\{42,43,44\}$.

Per-target errors reveal where the aggregate gains arise.
On Greenhouse, \frgr{} is lowest on indoor temperature, humidity,
and CO$_2$. TFT is lowest on PAR MAE;
\urgr{} is lowest on PAR RMSE.
On Wastewater, TFT is lowest on DO and PO$_4$ and on NH$_4$ MAE;
\flatrollgr{} is lowest on NH$_4$ RMSE. On VitalDB,
\frgr{} is lowest on BIS, a graph-covered drug-sensitive target, and
is comparable to \flatrollgr{} on MBP. 
TFT is lowest on HR MAE, while FR-GR is lowest on HR RMSE.
SpO$_2$ has no drug-effect edge in the prior
(Appendix~\ref{app:graph-prompt}), and TimesFM~2.5 attains the lowest MAE.
On PreDist, \frgr{} is lower than \fr{}, \flatrollgr{}, and \urgr{}
on net-return temperature and heat power, and lower than \fr{} and
\flatrollgr{} on flow. \urgr{} is lowest on supply temperature.
TFT is lowest on net-return temperature, and DAG
is lower on the algebraic heat-power and flow channels. This distinction is 
reflected in the role-specific construction of
Appendix~\ref{app:ie-head}: temperatures evolve as inertial states, while
heat power and flow are contemporaneous algebraic readouts
(Table~\ref{tab:ie-head}).

\begin{table*}[t]
  \centering
  \caption{Per-target test MAE / RMSE ($\downarrow$) on Greenhouse,
  physical units. Same methods as Table~\ref{tab:main}.}
  \label{tab:app-per-target-greenhouse}
  \scriptsize
  \setlength{\tabcolsep}{2.8pt}
  \resizebox{\textwidth}{!}{%
  \begin{tabular}{@{}lrrrrrrrr@{}}
    \toprule
    & \multicolumn{2}{c}{indoor $T$} & \multicolumn{2}{c}{RH} & \multicolumn{2}{c}{CO$_2$} & \multicolumn{2}{c}{PAR} \\
    \cmidrule(lr){2-3} \cmidrule(lr){4-5} \cmidrule(lr){6-7} \cmidrule(lr){8-9}
    Method & MAE & RMSE & MAE & RMSE & MAE & RMSE & MAE & RMSE \\
    \midrule
    Persistence
      & $1.489$ & $2.273$ & $5.51$ & $8.09$ & $81.5$ & $151.5$ & $73.8$ & $143.2$ \\
    Chronos-2
      & $1.618$ & $2.432$ & $5.68$ & $8.32$ & $78.0$ & $153.6$ & $71.3$ & $128.4$ \\
    Chronos-2-Cov
      & $1.450$ & $2.290$ & $5.18$ & $7.89$ & $76.7$ & $151.7$ & $67.7$ & $125.0$ \\
    TimesFM 2.5
      & $1.537$ & $2.317$ & $5.57$ & $8.08$ & $77.0$ & $148.6$ & $77.1$ & $138.1$ \\
    DAG
      & \pmstd{1.397}{0.026} & \pmstd{2.040}{0.035} & \pmstd{5.61}{0.08} & \pmstd{7.83}{0.08}
      & \pmstd{87.7}{0.5} & \pmstd{152.4}{3.1} & \pmstd{70.5}{1.8} & \pmstd{118.2}{2.4} \\
    TFT
      & \pmstd{1.199}{0.046} & \pmstd{1.720}{0.066} & \pmstd{4.75}{0.17} & \pmstd{6.43}{0.22}
      & \pmstd{75.7}{4.0} & \secondpm{119.9}{4.1} & \bestpm{60.3}{1.5} & \pmstd{102.5}{3.8} \\
    \midrule
    \ho{}
      & \pmstd{1.350}{0.014} & \pmstd{1.881}{0.028} & \pmstd{5.52}{0.15} & \pmstd{7.41}{0.16}
      & \pmstd{100.2}{1.4} & \pmstd{151.1}{1.9} & \pmstd{72.9}{0.6} & \pmstd{108.0}{1.1} \\
    \dc{}
      & \pmstd{1.283}{0.017} & \pmstd{1.747}{0.021} & \pmstd{5.10}{0.09} & \pmstd{6.93}{0.07}
      & \pmstd{88.2}{1.9} & \pmstd{131.2}{1.9} & \pmstd{72.1}{2.5} & \pmstd{106.9}{2.5} \\
    \sgraph{}
      & \pmstd{1.238}{0.024} & \pmstd{1.713}{0.031} & \pmstd{5.14}{0.05} & \pmstd{6.94}{0.02}
      & \pmstd{84.6}{2.3} & \pmstd{128.9}{2.1} & \pmstd{71.8}{2.4} & \pmstd{107.4}{1.0} \\
    \ie{}
      & \pmstd{1.256}{0.031} & \pmstd{1.729}{0.037} & \pmstd{5.20}{0.08} & \pmstd{7.02}{0.07}
      & \pmstd{85.9}{1.8} & \pmstd{129.5}{2.7} & \pmstd{72.2}{3.3} & \pmstd{108.1}{3.7} \\
    \rbel{}
      & \pmstd{1.265}{0.026} & \pmstd{1.744}{0.033} & \pmstd{5.17}{0.03} & \pmstd{6.95}{0.04}
      & \pmstd{84.2}{1.8} & \pmstd{127.9}{1.2} & \pmstd{70.2}{1.3} & \pmstd{106.9}{2.0} \\
    \fr{}
      & \pmstd{1.242}{0.021} & \pmstd{1.729}{0.013} & \secondpm{4.72}{0.13} & \secondpm{6.38}{0.16}
      & \pmstd{76.6}{1.1} & \pmstd{121.6}{1.0} & \pmstd{73.2}{2.6} & \pmstd{107.8}{3.0} \\
    \midrule
    \flatroll{}
      & \pmstd{1.194}{0.017} & \pmstd{1.669}{0.027} & \pmstd{4.89}{0.16} & \pmstd{6.59}{0.18}
      & \pmstd{82.4}{2.9} & \pmstd{127.2}{4.2} & \pmstd{68.0}{1.0} & \pmstd{103.2}{1.1} \\
    \flatrollgr{}
      & \pmstd{1.158}{0.027} & \pmstd{1.620}{0.028} & \pmstd{4.85}{0.05} & \pmstd{6.58}{0.07}
      & \pmstd{74.7}{1.4} & \secondpm{119.9}{4.3} & \pmstd{65.7}{1.4} & \pmstd{103.1}{1.8} \\
    \urgr{}
      & \secondpm{1.142}{0.013} & \secondpm{1.605}{0.027} & \pmstd{4.77}{0.04} & \pmstd{6.50}{0.06}
      & \secondpm{74.4}{1.8} & \pmstd{120.6}{2.9} & \pmstd{65.6}{1.8} & \bestpm{101.3}{2.3} \\
    \rbelgr{}
      & \pmstd{1.198}{0.014} & \pmstd{1.661}{0.016} & \pmstd{5.18}{0.13} & \pmstd{7.04}{0.14}
      & \pmstd{77.6}{1.4} & \pmstd{123.9}{0.2} & \pmstd{67.2}{1.9} & \pmstd{104.0}{1.9} \\
    \frgr{}
      & \bestpm{1.134}{0.025} & \bestpm{1.588}{0.047} & \bestpm{4.59}{0.05} & \bestpm{6.29}{0.08}
      & \bestpm{68.8}{1.2} & \bestpm{112.8}{2.6} & \secondpm{65.1}{0.6} & \secondpm{102.2}{1.7} \\
    \bottomrule
  \end{tabular}}
\end{table*}

\begin{table*}[t]
  \centering
  \caption{Per-target test MAE / RMSE ($\downarrow$) on Wastewater,
  physical units. Same methods as Table~\ref{tab:main}.}
  \label{tab:app-per-target-wastewater}
  \scriptsize
  \setlength{\tabcolsep}{3.6pt}
  \resizebox{\textwidth}{!}{%
  \begin{tabular}{@{}lrrrrrr@{}}
    \toprule
    & \multicolumn{2}{c}{DO} & \multicolumn{2}{c}{NH$_4$} & \multicolumn{2}{c}{PO$_4$} \\
    \cmidrule(lr){2-3} \cmidrule(lr){4-5} \cmidrule(lr){6-7}
    Method & MAE & RMSE & MAE & RMSE & MAE & RMSE \\
    \midrule
    Persistence
      & $0.813$ & $1.223$ & $0.768$ & $1.105$ & $0.549$ & $0.825$ \\
    Chronos-2
      & $0.604$ & $0.988$ & $0.693$ & $1.066$ & $0.459$ & $0.712$ \\
    Chronos-2-Cov
      & $0.484$ & $0.828$ & $0.474$ & $0.785$ & $0.338$ & $0.572$ \\
    TimesFM 2.5
      & $0.567$ & $0.920$ & $0.664$ & $0.995$ & $0.439$ & $0.668$ \\
    DAG
      & \pmstd{0.533}{0.011} & \pmstd{0.830}{0.009} & \pmstd{0.431}{0.011}
      & \pmstd{0.688}{0.012} & \pmstd{0.307}{0.006} & \pmstd{0.492}{0.008} \\
    TFT
      & \bestpm{0.338}{0.002} & \bestpm{0.574}{0.007} & \bestpm{0.320}{0.015}
      & \pmstd{0.574}{0.037} & \bestpm{0.222}{0.002} & \bestpm{0.371}{0.003} \\
    \midrule
    \ho{}
      & \pmstd{0.589}{0.011} & \pmstd{0.855}{0.012} & \pmstd{0.630}{0.008}
      & \pmstd{0.898}{0.014} & \pmstd{0.423}{0.004} & \pmstd{0.607}{0.001} \\
    \dc{}
      & \pmstd{0.403}{0.001} & \pmstd{0.648}{0.007} & \pmstd{0.425}{0.012}
      & \pmstd{0.651}{0.017} & \pmstd{0.296}{0.014} & \pmstd{0.468}{0.016} \\
    \sgraph{}
      & \pmstd{0.389}{0.008} & \pmstd{0.628}{0.007} & \pmstd{0.413}{0.017}
      & \pmstd{0.650}{0.028} & \pmstd{0.281}{0.007} & \pmstd{0.450}{0.005} \\
    \ie{}
      & \pmstd{0.395}{0.001} & \pmstd{0.636}{0.004} & \pmstd{0.415}{0.021}
      & \pmstd{0.659}{0.027} & \pmstd{0.285}{0.009} & \pmstd{0.451}{0.014} \\
    \rbel{}
      & \pmstd{0.394}{0.005} & \pmstd{0.633}{0.006} & \pmstd{0.397}{0.005}
      & \pmstd{0.626}{0.005} & \pmstd{0.273}{0.004} & \pmstd{0.434}{0.003} \\
    \fr{}
      & \pmstd{0.363}{0.001} & \pmstd{0.598}{0.003} & \pmstd{0.351}{0.013}
      & \pmstd{0.573}{0.020} & \pmstd{0.244}{0.004} & \pmstd{0.396}{0.008} \\
    \midrule
    \flatroll{}
      & \pmstd{0.366}{0.007} & \pmstd{0.594}{0.007} & \pmstd{0.353}{0.024}
      & \pmstd{0.570}{0.028} & \pmstd{0.240}{0.006} & \pmstd{0.388}{0.009} \\
    \flatrollgr{}
      & \secondpm{0.350}{0.014} & \pmstd{0.584}{0.017} & \pmstd{0.330}{0.002}
      & \bestpm{0.529}{0.011} & \pmstd{0.235}{0.002} & \secondpm{0.384}{0.003} \\
    \urgr{}
      & \pmstd{0.358}{0.010} & \pmstd{0.594}{0.011} & \pmstd{0.345}{0.002}
      & \pmstd{0.559}{0.008} & \pmstd{0.245}{0.004} & \pmstd{0.393}{0.004} \\
    \rbelgr{}
      & \pmstd{0.363}{0.008} & \pmstd{0.600}{0.009} & \pmstd{0.356}{0.004}
      & \pmstd{0.567}{0.007} & \pmstd{0.254}{0.000} & \pmstd{0.409}{0.006} \\
    \frgr{}
      & \pmstd{0.353}{0.009} & \secondpm{0.582}{0.012} & \secondpm{0.326}{0.006}
      & \secondpm{0.537}{0.017} & \secondpm{0.231}{0.003} & \secondpm{0.384}{0.005} \\
    \bottomrule
  \end{tabular}}
\end{table*}

\begin{table*}[t]
  \centering
  \caption{Per-target test MAE / RMSE ($\downarrow$) on VitalDB,
  physical units. Same methods as Table~\ref{tab:main}.}
  \label{tab:app-per-target-vitaldb}
  \scriptsize
  \setlength{\tabcolsep}{2.8pt}
  \resizebox{\textwidth}{!}{%
  \begin{tabular}{@{}lrrrrrrrr@{}}
    \toprule
    & \multicolumn{2}{c}{MBP} & \multicolumn{2}{c}{HR} & \multicolumn{2}{c}{SpO$_2$} & \multicolumn{2}{c}{BIS} \\
    \cmidrule(lr){2-3} \cmidrule(lr){4-5} \cmidrule(lr){6-7} \cmidrule(lr){8-9}
    Method & MAE & RMSE & MAE & RMSE & MAE & RMSE & MAE & RMSE \\
    \midrule
    Persistence
      & $5.948$ & $10.745$ & $3.714$ & $7.126$ & $0.235$ & $1.050$ & $4.892$ & $7.087$ \\
    Chronos-2
      & $6.063$ & $10.455$ & $3.470$ & $6.416$ & $\underline{0.229}$ & $1.025$ & $4.693$ & $6.995$ \\
    Chronos-2-Cov
      & $5.992$ & $10.463$ & $3.533$ & $6.648$ & $0.230$ & $1.036$ & $4.676$ & $7.035$ \\
    TimesFM 2.5
      & $5.671$ & $9.679$ & $\underline{3.334}$ & $6.205$ & $\mathbf{0.214}$ & $0.952$ & $4.564$ & $6.800$ \\
    DAG
      & \pmstd{5.608}{0.020} & \pmstd{9.335}{0.032} & \pmstd{3.526}{0.017} & \pmstd{6.318}{0.026}
      & \pmstd{0.280}{0.003} & \pmstd{1.050}{0.004} & \pmstd{4.638}{0.011} & \pmstd{6.864}{0.025} \\
    TFT
      & \pmstd{5.117}{0.032} & \pmstd{8.685}{0.050} & \bestpm{3.294}{0.040} & \pmstd{5.959}{0.038}
      & \pmstd{0.350}{0.109} & \pmstd{1.139}{0.186} & \pmstd{4.140}{0.017} & \pmstd{5.867}{0.034} \\
    \midrule
    \ho{}
      & \pmstd{5.401}{0.050} & \pmstd{8.867}{0.006} & \pmstd{3.401}{0.018} & \pmstd{5.885}{0.003}
      & \pmstd{0.282}{0.004} & \pmstd{0.896}{0.002} & \pmstd{4.441}{0.007} & \pmstd{6.442}{0.008} \\
    \dc{}
      & \pmstd{5.308}{0.062} & \pmstd{8.675}{0.014} & \pmstd{3.457}{0.056} & \pmstd{5.873}{0.027}
      & \pmstd{0.298}{0.006} & \pmstd{0.889}{0.004} & \pmstd{4.217}{0.007} & \pmstd{5.926}{0.009} \\
    \sgraph{}
      & \pmstd{5.335}{0.041} & \pmstd{8.696}{0.036} & \pmstd{3.416}{0.052} & \pmstd{5.873}{0.007}
      & \pmstd{0.341}{0.016} & \pmstd{0.899}{0.001} & \pmstd{4.232}{0.022} & \pmstd{5.944}{0.012} \\
    \ie{}
      & \pmstd{5.290}{0.018} & \pmstd{8.670}{0.022} & \pmstd{3.417}{0.052} & \pmstd{5.880}{0.013}
      & \pmstd{0.333}{0.017} & \pmstd{0.897}{0.003} & \pmstd{4.211}{0.009} & \pmstd{5.938}{0.010} \\
    \rbel{}
      & \pmstd{5.276}{0.016} & \pmstd{8.657}{0.034} & \pmstd{3.393}{0.003} & \pmstd{5.875}{0.007}
      & \pmstd{0.348}{0.011} & \pmstd{0.901}{0.004} & \pmstd{4.122}{0.020} & \pmstd{5.809}{0.032} \\
    \fr{}
      & \pmstd{5.168}{0.027} & \pmstd{8.549}{0.017} & \pmstd{3.391}{0.023} & \pmstd{5.871}{0.016}
      & \pmstd{0.331}{0.017} & \pmstd{0.898}{0.001} & \pmstd{4.087}{0.013} & \pmstd{5.749}{0.019} \\
    \midrule
    \flatroll{}
      & \pmstd{5.176}{0.027} & \pmstd{8.501}{0.011} & \pmstd{3.391}{0.012} & \pmstd{5.881}{0.015}
      & \pmstd{0.337}{0.004} & \pmstd{0.895}{0.004} & \pmstd{4.086}{0.009} & \secondpm{5.723}{0.007} \\
    \flatrollgr{}
      & \bestpm{5.097}{0.027} & \pmstd{8.497}{0.011} & \pmstd{3.345}{0.003} & \pmstd{5.844}{0.015}
      & \pmstd{0.316}{0.010} & \bestpm{0.886}{0.005} & \pmstd{4.089}{0.018} & \pmstd{5.735}{0.022} \\
    \urgr{}
      & \pmstd{5.128}{0.041} & \secondpm{8.495}{0.017} & \pmstd{3.350}{0.011} & \pmstd{5.850}{0.011}
      & \pmstd{0.317}{0.013} & \pmstd{0.890}{0.001} & \pmstd{4.103}{0.008} & \pmstd{5.744}{0.012} \\
    \rbelgr{}
      & \pmstd{5.148}{0.035} & \pmstd{8.542}{0.009} & \pmstd{3.350}{0.005} & \secondpm{5.843}{0.005}
      & \pmstd{0.320}{0.015} & \secondpm{0.888}{0.001} & \secondpm{4.084}{0.028} & \pmstd{5.732}{0.028} \\
    \frgr{}
      & \secondpm{5.103}{0.031} & \bestpm{8.487}{0.006} & \pmstd{3.347}{0.019} & \bestpm{5.831}{0.010}
      & \pmstd{0.319}{0.023} & \secondpm{0.888}{0.003} & \bestpm{4.071}{0.014} & \bestpm{5.715}{0.017} \\
    \bottomrule
  \end{tabular}}
\end{table*}

\begin{table*}[t]
  \centering
  \caption{Per-target test MAE / RMSE ($\downarrow$) on PreDist$^{\dagger}$,
  physical units. Same methods as Table~\ref{tab:main}.}
  \label{tab:app-per-target-predist}
  \scriptsize
  \setlength{\tabcolsep}{2.6pt}
  \resizebox{\textwidth}{!}{%
  \begin{tabular}{@{}lrrrrrrrrrr@{}}
    \toprule
    & \multicolumn{2}{c}{Supply $T$} & \multicolumn{2}{c}{HC ret.\ $T$}
      & \multicolumn{2}{c}{Net ret.\ $T$} & \multicolumn{2}{c}{Heat power}
      & \multicolumn{2}{c}{Flow} \\
    \cmidrule(lr){2-3} \cmidrule(lr){4-5} \cmidrule(lr){6-7}
    \cmidrule(lr){8-9} \cmidrule(lr){10-11}
    Method & MAE & RMSE & MAE & RMSE & MAE & RMSE & MAE & RMSE & MAE & RMSE \\
    \midrule
    Persistence
      & $2.641$ & $4.164$ & $2.470$ & $3.732$ & $2.450$ & $3.958$
      & $21.85$ & $37.40$ & $0.617$ & $1.064$ \\
    Chronos-2
      & $2.295$ & $3.740$ & $2.163$ & $3.427$ & $2.144$ & $3.504$
      & $19.95$ & $34.15$ & $0.559$ & $0.954$ \\
    Chronos-2-Cov
      & $1.772$ & $3.061$ & $2.168$ & $3.521$ & $2.117$ & $3.630$
      & $\underline{18.83}$ & $35.93$ & $\underline{0.524}$ & $0.984$ \\
    TimesFM 2.5
      & $2.182$ & $3.591$ & $2.024$ & $3.205$ & $2.014$ & $3.313$
      & $18.84$ & $32.18$ & $0.527$ & $\underline{0.904}$ \\
    DAG
      & \pmstd{1.671}{0.022} & \pmstd{2.588}{0.058} & \pmstd{1.938}{0.025} & \pmstd{2.940}{0.045}
      & \pmstd{1.969}{0.030} & \pmstd{3.107}{0.045} & \bestpm{17.60}{0.14} & \bestpm{30.30}{0.28}
      & \bestpm{0.492}{0.003} & \bestpm{0.840}{0.006} \\
    TFT
      & \pmstd{1.455}{0.089} & \pmstd{2.075}{0.103} & \pmstd{1.950}{0.113} & \pmstd{2.816}{0.117}
      & \bestpm{1.873}{0.036} & \bestpm{2.852}{0.044} & \pmstd{22.98}{2.83} & \pmstd{38.29}{4.60}
      & \pmstd{0.872}{0.243} & \pmstd{1.344}{0.305} \\
    \midrule
    \ho{}
      & \pmstd{2.191}{0.073} & \pmstd{3.271}{0.013} & \pmstd{2.088}{0.036} & \pmstd{3.118}{0.012}
      & \pmstd{2.277}{0.016} & \pmstd{3.360}{0.016} & \pmstd{22.08}{0.76} & \pmstd{34.79}{0.82}
      & \pmstd{0.676}{0.068} & \pmstd{1.083}{0.094} \\
    \dc{}
      & \pmstd{1.598}{0.041} & \pmstd{2.391}{0.042} & \pmstd{1.977}{0.032} & \pmstd{2.925}{0.027}
      & \pmstd{2.139}{0.083} & \pmstd{3.195}{0.088} & \pmstd{22.88}{1.71} & \pmstd{35.49}{2.43}
      & \pmstd{0.712}{0.028} & \pmstd{1.131}{0.055} \\
    \sgraph{}
      & \pmstd{1.595}{0.015} & \pmstd{2.371}{0.043} & \pmstd{1.965}{0.021} & \pmstd{2.937}{0.012}
      & \pmstd{2.054}{0.076} & \pmstd{3.124}{0.111} & \pmstd{22.55}{2.48} & \pmstd{35.66}{3.61}
      & \pmstd{0.713}{0.054} & \pmstd{1.128}{0.074} \\
    \ie{}
      & \pmstd{1.665}{0.045} & \pmstd{2.475}{0.050} & \pmstd{2.046}{0.008} & \pmstd{3.018}{0.021}
      & \pmstd{2.159}{0.054} & \pmstd{3.218}{0.027} & \pmstd{22.60}{1.51} & \pmstd{35.92}{1.96}
      & \pmstd{0.658}{0.062} & \pmstd{1.052}{0.083} \\
    \rbel{}
      & \pmstd{1.530}{0.016} & \pmstd{2.338}{0.065} & \pmstd{1.967}{0.012} & \pmstd{2.932}{0.022}
      & \pmstd{2.022}{0.028} & \pmstd{3.085}{0.025} & \pmstd{22.15}{2.10} & \pmstd{35.30}{2.97}
      & \pmstd{0.643}{0.047} & \pmstd{1.033}{0.076} \\
    \fr{}
      & \pmstd{1.290}{0.040} & \pmstd{1.902}{0.029} & \bestpm{1.817}{0.036} & \secondpm{2.602}{0.028}
      & \pmstd{1.980}{0.058} & \pmstd{2.921}{0.051} & \pmstd{23.28}{3.03} & \pmstd{36.68}{4.26}
      & \pmstd{0.689}{0.082} & \pmstd{1.074}{0.105} \\
    \midrule
    \flatroll{}
      & \pmstd{1.310}{0.047} & \pmstd{1.909}{0.010} & \pmstd{1.917}{0.102} & \pmstd{2.653}{0.071}
      & \pmstd{2.209}{0.209} & \pmstd{3.107}{0.162} & \pmstd{23.50}{1.05} & \pmstd{37.33}{1.62}
      & \pmstd{0.699}{0.112} & \pmstd{1.094}{0.163} \\
    \flatrollgr{}
      & \secondpm{1.284}{0.022} & \secondpm{1.879}{0.027} & \secondpm{1.820}{0.066} & \bestpm{2.561}{0.035}
      & \pmstd{2.050}{0.239} & \pmstd{2.936}{0.186} & \pmstd{20.04}{1.59} & \pmstd{32.28}{1.74}
      & \pmstd{0.583}{0.040} & \pmstd{0.944}{0.057} \\
    \urgr{}
      & \bestpm{1.252}{0.008} & \bestpm{1.843}{0.014} & \pmstd{1.871}{0.046} & \pmstd{2.622}{0.033}
      & \pmstd{2.036}{0.078} & \pmstd{2.949}{0.063} & \pmstd{19.60}{0.79} & \pmstd{31.99}{1.11}
      & \pmstd{0.561}{0.035} & \pmstd{0.923}{0.050} \\
    \rbelgr{}
      & \pmstd{1.496}{0.008} & \pmstd{2.168}{0.058} & \pmstd{2.040}{0.154} & \pmstd{2.960}{0.177}
      & \pmstd{1.967}{0.038} & \pmstd{3.031}{0.023} & \pmstd{20.26}{0.90} & \pmstd{32.53}{1.06}
      & \pmstd{0.592}{0.015} & \pmstd{0.965}{0.024} \\
    \frgr{}
      & \pmstd{1.290}{0.030} & \pmstd{1.905}{0.024} & \pmstd{1.834}{0.006} & \pmstd{2.616}{0.008}
      & \secondpm{1.928}{0.069} & \secondpm{2.871}{0.042} & \pmstd{18.85}{0.58} & \secondpm{30.88}{0.52}
      & \pmstd{0.562}{0.021} & \pmstd{0.926}{0.028} \\
    \bottomrule
  \end{tabular}}
\end{table*}

\section{Instantaneous-effect head}
\label{app:ie-head}

The instantaneous-effect head is an extension for contemporaneous
algebraic targets, used on PreDist.

An observation vector can mix two constitutive roles. Delayed inertial
targets evolve as states with memory. Instantaneous algebraic targets are
contemporaneous functions of those states and of the current drivers. A
single JEPA latent $z_{t+h}$ then packs both roles into one vector. The
teacher, the predictor, and the decoder share one interface for a slow
state and a same-horizon readout.

The instantaneous-effect head splits that interface by role
$c\in\{\mathrm{dyn},\mathrm{alg}\}$. Inertial channels keep family-effect
memory through Eq.~(\ref{eq:effect-dyn}). Algebraic channels are evaluated
on the same horizon and never enter $\mathcal{F}$. The construction follows the
target property: delayed inertial channels together with instantaneous
algebraic channels.

PreDist is the evaluation system with this mix. Supply and return
temperatures store heat and lag valve, setpoint, and pump changes. Heat
power and volume flow are algebraic functions of those temperatures and of
the hydraulic drivers at the same $10$\,min step. The typed prior follows
the same split (Appendix~\ref{app:graph-prompt}): $G_{\mathrm{dyn}}$ routes
driver$\rightarrow$effect$\rightarrow$temperature;
$G_{\mathrm{alg}}$ routes valve/pump/three-way$\rightarrow$flow and
flow$+\Delta T\rightarrow$heat power on the same horizon.

\textbf{Independent observation head.}
The context encoder remains joint over the history window. The head
indexes the EMA teacher, the predictor queries, and the decoder by
role. Each role keeps its own JEPA latent, and each decoder head
reconstructs only that role's channels.
A shared-latent baseline that packs both roles into one $z_{t+h}$,
treats algebraic channels as additional inertial states, and decodes
every channel from that vector is the comparison in
Table~\ref{tab:ie-head}.

\textbf{Family persistence $\lambda_f$.}
In Eq.~(\ref{eq:effect-dyn}), $\lambda_f=\sigma(\rho_f)$ is the learned
retention of a \emph{latent-effect family}. It says how long a driver
injection remains in $z_{f,\tau}$. That persistence is the right object
for delayed influence: a valve or setpoint command continues to move
inertial temperatures after it is issued. Instantaneous algebraic
targets have no such reservoir. Their value at $t+h$ is fixed once
contemporaneous inertial states and drivers are known.

The \ie{} ablation removes the $\lambda_f z_{f,\tau}$ term, so an effect
node does not carry driver history. An algebraic observation is not an
effect node with $\lambda_f=0$. It is a constraint on the current inertial
state and drivers, and it never enters $Z^{F}$. Family persistence
governs how long those drivers remain available to the inertial targets.
The instantaneous-effect head governs how the two kinds of
\emph{observation} are represented in JEPA space.

\textbf{Construction.}
Let $I_{\mathrm{dyn}}$ index the inertial channels and $I_{\mathrm{alg}}$
the algebraic channels. On PreDist these are the three temperatures and
the pair (heat power, flow). The context encoder $E_\theta$ remains joint
over the full history. The EMA teacher is applied once per role to the
same future sequence with a causal mask and complementary channels
zeroed:
\begin{equation}
\label{eq:role-teacher}
  z_{t+1:t+H}^{*,c}
  =
  \operatorname{sg}\!\bigl(E_{\bar\theta}(X_{t+1:t+H}^{c})\bigr),
  \qquad
  c\in\{\mathrm{dyn},\mathrm{alg}\},
\end{equation}
where $X_{t+1:t+H}^{c}$ equals $X_{t+1:t+H}$ on $I_c$ and is zero
elsewhere, and $z_{t+h}^{*,c}$ depends only on $X_{t+1:t+h}^{c}$.
A shared predictor, and under \gr{} a shared residual adapter, emit
$\hat z_{t+h}^{c}$ from a role query $r_c$. Role identity is that query
embedding; driver tokens are shared across roles. When a
dual-timescale prior is present, dyn queries attend the $G_{\mathrm{dyn}}$
readout with algebraic nodes zeroed, and alg queries attend the merged
$G_{\mathrm{alg}}$ readout.
At horizon $h$, $G_{\mathrm{dyn}}$ is evaluated from the rolled family
states $Z_{t+h}^{F}$. $G_{\mathrm{alg}}$ then overwrites the
action and exogenous nodes with the encodings of $D_{t+h-1}$ and
message-passes on algebraic edges.
The heating-circuit return parent of heat power uses that
$G_{\mathrm{dyn}}$ embedding as its source: static identity and type,
updated by effect-to-temperature messages from those rolled states
(Appendix~\ref{app:arch}).
Decoder heads map $\hat z_{t+h}^{\mathrm{dyn}}$
onto $I_{\mathrm{dyn}}$ and $\hat z_{t+h}^{\mathrm{alg}}$ onto
$I_{\mathrm{alg}}$. The JEPA term averages the two role residuals. The same
driver--graph--rollout--residual stack of Section~\ref{sec:method} is used
throughout. \ho{} and \dc{} keep the role split without a GNN, reading
roles from the channel inventory.

Table~\ref{tab:ie-head} compares that head to a shared-latent
\frgr{} that packs all five PreDist targets into one $z_{t+h}$.
Removing the head raises pooled nMAE from $0.0495$ to $0.0582$,
above DAG ($0.0512$) and TimesFM~2.5 ($0.0563$) in
Table~\ref{tab:main}. The increase is concentrated on the algebraic
channels: heat-power MAE rises from $18.85$ to $28.52$ and flow from
$0.562$ to $0.730$. The three inertial temperatures remain close;
heating-circuit return is unchanged. The current shared-latent packing
is therefore a worse interface for this mix. The head keeps delayed
temperatures and contemporaneous readouts in separate JEPA spaces, and
that split recovers the algebraic maps.

\begin{table}[ht]
  \centering
  \caption{PreDist test error of \frgr{} with a shared JEPA latent
  versus with the instantaneous-effect head. Pooled nMAE on $[-1,1]$;
  per-target MAE in physical units. Mean $\pm$ sample std over seeds
  $\{42,43,44\}$. The headed row is the PreDist \frgr{} entry of
  Tables~\ref{tab:main} and~\ref{tab:app-per-target-predist}.}
  \label{tab:ie-head}
  \small
  \setlength{\tabcolsep}{3.0pt}
  \resizebox{\columnwidth}{!}{%
  \begin{tabular}{@{}lrrrrrr@{}}
    \toprule
    & nMAE & Supply $T$ & HC ret.\ $T$ & Net ret.\ $T$
      & Heat & Flow \\
    \midrule
    \frgr{}, shared $z$
      & \pmstd{0.0582}{0.0020} & \pmstd{1.447}{0.078}
      & \pmstd{1.823}{0.073} & \pmstd{2.104}{0.138}
      & \pmstd{28.52}{3.13} & \pmstd{0.730}{0.049} \\
    \frgr{}
      & \bestpm{0.0495}{0.0003} & \bestpm{1.290}{0.030}
      & \pmstd{1.834}{0.006} & \bestpm{1.928}{0.069}
      & \bestpm{18.85}{0.58} & \bestpm{0.562}{0.021} \\
    \bottomrule
  \end{tabular}}
\end{table}

\section{Architecture details}
\label{app:arch}

This appendix records the computational interfaces of
Sections~\ref{sec:problem}--\ref{sec:method}.

\textbf{EMA teacher.}
$E_{\bar\theta}$ is a momentum-updated copy of $E_\theta$ with the same
architecture.
After each optimizer step, every parameter is updated by EMA with
decay $0.996$.
$E_\theta$ and $E_{\bar\theta}$ use the same learned positional
embeddings, evaluated at indices $0,\ldots,L{-}1$ of their respective
input sequences.
During training, $E_{\bar\theta}$ encodes the ground-truth future
sequence $X_{t+1:t+H}$ with a causal mask.
Its $h$-th output token defines the stop-gradient target $z_{t+h}^{*}$
and depends only on $X_{t+1:t+h}$.
On PreDist, the teacher is applied once per role to the same future
sequence with complementary channels zeroed
(Appendix~\ref{app:ie-head}).

\textbf{Non-effect node features.}
Every node is initialized as
$e_i^{\mathrm{static}}=e_i^{\mathrm{id}}+e_i^{\mathrm{type}}$.
Effect nodes then receive the gated injection of $z_{f,\tau}$
(Eq.~\ref{eq:inject}).
Exogenous nodes are source-only in the prior, so reverse edges into
them are omitted; once written, their input encodings are restored
after each GNN layer.
Target nodes remain at their static embeddings until message passing.
The GNN is re-evaluated separately at each horizon using
$\{Z_{t+h}^{F}\}_{h=1}^{H}$ and $\{D_{t+h-1}\}_{h=1}^{H}$.

On a single-timescale prior, action and exogenous nodes are overwritten
with their corresponding encodings from $D_{t+h-1}$, the last driver
of the rollout prefix, before message passing.
Horizon-$h$ node features therefore comprise static identity and type,
rolled family states $Z_{t+h}^{F}$, and those driver encodings.
On a dual-timescale prior, $G_{\mathrm{dyn}}$ is first evaluated from
$Z_{t+h}^{F}$ without this horizon-driver overwrite.
$G_{\mathrm{alg}}$ then overwrites only action and exogenous nodes
with their corresponding encodings from $D_{t+h-1}$ and
message-passes on algebraic edges.
On PreDist, the heating-circuit return parent of heat power is read
from this $G_{\mathrm{dyn}}$ embedding.

\textbf{GNN edge attributes.}
Relation labels become discrete edge types with a learned matrix
$W_r$. Confidence initializes a learnable scalar gate
$g=\sigma(\ell)$, with $\ell=\operatorname{logit}(\operatorname{clip}(c,\varepsilon,1-\varepsilon))$.
Messages are $g\,W_r h_{\mathrm{src}}$. Sign is stored as metadata in
the prior.

\textbf{Predictor attention.}
Query $h$ conditions on $H_{t+h}^{G}$ by cross-attention to that
horizon's graph nodes.
Cross-attention to encoded driver tokens is prefix-masked: query $h$
attends $D_{t:t+h-1}$.
Under \gr{}, those tokens enter only through $A_R$, with the same
prefix mask.
Horizon queries then share decoder self-attention, with no causal
mask among queries.
The recurrence is sequential; the decoder is joint over the
known-action horizon.

\textbf{Residual adapter.}
$A_R$ is a Transformer decoder over encoded tokens of $D_{t:t+H-1}$
and history $H_t^{X}$. Cross-attention uses the same prefix mask;
query self-attention is unmasked.
Its last linear layer is zero-initialized.

\section{Training configuration}
\label{app:train}

Tables~\ref{tab:app-train-shared}--\ref{tab:app-train-data} record the
\frgr{} recipe used in Table~\ref{tab:main}. Wastewater uses $50$
epochs rather than $30$ because validation nMAE continues to decrease
through the longer schedule; the remaining datasets use $30$ epochs.
The latent-effect window
equals the context length $L_X$ (Table~\ref{tab:app-multi-windows}).
VitalDB uses a matched-window control: windows are indexed as if
$L_A=180$, then past drivers are truncated to $L_X=90$. Per-family
timescales $\tau_f$ initialize $\lambda_f$ in steps: VitalDB propofol 24,
remifentanil 12; Greenhouse thermal 24, ventilation 6, light 2, screen 4,
CO$_2$ 6; Wastewater dosing 8, aeration 6, phase 15; PreDist valve 6,
setpoint 18, pump 1. A shared $\tau=8$ is compared in
Appendix~\ref{app:hyper}. The glucose protocols in
Appendix~\ref{app:single-target} use the same \frgr{} backbone with
batch size 16 and learning rate $3{\times}10^{-4}$; Shanghai uses a
typed event encoder and CGMacros a meal encoder.

\begin{table}[ht]
  \centering
  \caption{Shared \frgr{} architecture and optimization.
  PreDist uses the large-batch row of Table~\ref{tab:app-train-data};
  Wastewater uses the 50-epoch row.}
  \label{tab:app-train-shared}
  \footnotesize
  \setlength{\tabcolsep}{4pt}
  \begin{tabular}{@{}ll@{}}
    \toprule
    Setting & Value \\
    \midrule
    Width $d$ / heads / FFN
      & $128$ / $4$ / $4d$ \\
    History encoder
      & 2-layer causal Transformer, Pre-LN, GELU \\
    GNN / predictor / residual adapter
      & 2 relation-aware layers; 2-layer Transformer decoder \\
    Dropout / positional encoding
      & $0.1$; learned embeddings on the state encoder;
        sinusoidal $\Delta t$ on driver events \\
    EMA target decay
      & $0.996$ \\
    Optimizer
      & AdamW, weight decay $10^{-4}$, gradient clip $1.0$ \\
    Schedule
      & cosine, 30 epochs (50 on Wastewater) \\
    Default learning rate
      & $3{\times}10^{-4}$, min $10^{-5}$, no warmup \\
    Loss weights
      & $\lambda_G=0.1$,
        $\lambda_{\mathrm{cap}}=0.01$,
        $\lambda_{\mathrm{obs}}=1$ \\
    Residual scale
      & $\alpha^{R}_{\max}=0.5$, init $0.1$;
        graph-gate logit $-4$ \\
    Seeds / hardware
      & $\{42,43,44\}$; one NVIDIA A100 \\
    \bottomrule
  \end{tabular}
\end{table}

\begin{table}[ht]
  \centering
  \caption{Dataset-specific \frgr{} training settings.
  Train windows are the indexed training set.
  Wall-clock is one seed, train and validation.
  Wastewater uses $50$ epochs, batch $32$, learning rate
  $1.2{\times}10^{-3}$, min LR $4{\times}10^{-5}$, and a 3-epoch
  warmup; remaining datasets use $30$ epochs.
  PreDist uses a larger batch, learning rate $10^{-3}$, and min
  LR $3{\times}10^{-5}$.}
  \label{tab:app-train-data}
  \footnotesize
  \setlength{\tabcolsep}{3.2pt}
  \begin{tabular}{@{}lrrrrrrr@{}}
    \toprule
    Dataset & Batch & LR & LR warm. & Stride & Train $N$ & Params & Time \\
    \midrule
    VitalDB
      & 8 & $3{\times}10^{-4}$ & 0 & 30 & 45{,}425
      & 2.0M & 4.0\,h \\
    Greenhouse
      & 8 & $3{\times}10^{-4}$ & 0 & 36 & 2{,}144
      & 2.9M & 0.4\,h \\
    Wastewater
      & 32 & $1.2{\times}10^{-3}$ & 3 & 90 & 4{,}093
      & 2.9M & 0.4\,h \\
    PreDist
      & 64 & $1{\times}10^{-3}$ & 1 & 36 & 87{,}475
      & 3.1M & 2.1\,h \\
    \bottomrule
  \end{tabular}
\end{table}

Parameter counts stay near $2$--$3$M. One run finishes in $0.4$--$4$
hours on a single A100, with VitalDB taking the longest. Latent recurrence is sequential in $L_X{+}H$ and linear
in $|V_F|$. The $H$ GNN evaluations are independent given the rolled states
and horizon-specific driver encodings, and are batched over the horizon. The history encoder
and predictor dominate wall-clock cost.

\section{Future-action perturbation protocol}
\label{app:perturb}

Section~\ref{sec:results-perturb} offsets continuous future-action
channels on the second half of the horizon, $I=[H/2,H)$, and reports
the relative nMAE increase $\Delta_{\mathrm{perturb}}$ on the full
window. The offset is applied in the normalized action space used at
training. Categorical actions and exogenous inputs are held fixed.
Continuous channels admit a shared additive offset of controlled
magnitude. Categorical substitutions have no common scale: a label
change can be nearly inert or can switch the operating mode.
The diagnostic therefore tests whether forecasts use later
continuous-action inputs.

\begin{table}[ht]
  \centering
  \caption{Second-half continuous-action offset.
  $\delta$ is the per-channel magnitude; values are clipped to the
  listed range.}
  \label{tab:app-perturb}
  \footnotesize
  \setlength{\tabcolsep}{4pt}
  \begin{tabular}{lllll}
    \toprule
    Dataset & $H$ & Interval $I$ & Range & $\delta$ \\
    \midrule
    VitalDB & $36$ ($6$\,min) & $[18,36)$ ($3$\,min) & $[0,1]$ & $0.15$ \\
    Greenhouse & $36$ ($3$\,h) & $[18,36)$ ($90$\,min) & $[0,1]$ & $0.15$ \\
    Wastewater & $90$ ($3$\,h) & $[45,90)$ ($90$\,min) & $[-1,1]$ & $0.30$ \\
    PreDist & $36$ ($6$\,h) & $[18,36)$ ($3$\,h) & $[-1,1]$ & $0.30$ \\
    \bottomrule
  \end{tabular}
\end{table}

Per-channel signs $s_j\in\{+1,-1\}$ are drawn once 
and shared across methods. If more than $10\%$ of steps
in $I$ clip for a sample-channel, that sample-channel sign is
flipped. Clipped values remain in the training range.

\section{Residual scale and reliance}
\label{app:residual}

Table~\ref{tab:residual} reports the converged residual scale
$\alpha^{R}$ and realized residual reliance $\rho_{R}$ of \frgr{}.
Across datasets, $\rho_{R}$ is $1.4$--$3.9\%$: the residual has small
relative latent magnitude compared with the graph base. Removing it at
evaluation (\frgrgo{} in Table~\ref{tab:routing}) nevertheless raises
nMAE and nRMSE on every dataset; the larger degradation on Greenhouse
and Wastewater coincides with their larger $\rho_{R}$. A small
residual is sufficient to recover information omitted by the graph
abstraction. Sensitivity of the same quantities
to residual-scale initialization and graph-base supervision is in
Appendix~\ref{app:hyper}.

\begin{table}[ht]
  \centering
  \caption{Converged residual scale $\alpha^{R}$ and reliance
  $\rho_{R}$ (Eq.~\ref{eq:rho}) of \frgr{}.}
  \label{tab:residual}
  \small
  \begin{tabular}{@{}lrr@{}}
    \toprule
    Dataset & $\alpha^{R}$ & $\rho_{R}$ \\
    \midrule
    Greenhouse
      & \pmstd{0.107}{0.001} & \pmstd{0.0302}{0.0019} \\
    Wastewater
      & \pmstd{0.104}{0.001} & \pmstd{0.0391}{0.0032} \\
    VitalDB
      & \pmstd{0.050}{0.002} & \pmstd{0.0142}{0.0009} \\
    PreDist
      & \pmstd{0.069}{0.001} & \pmstd{0.0143}{0.0016} \\
    \bottomrule
  \end{tabular}
\end{table}

\section{Hyperparameter sensitivity}
\label{app:hyper}

We evaluate the sensitivity of \frgr{} to four design choices across
the four multi-target protocols: residual-scale initialization,
graph-base supervision weight, latent-effect history length, and
family-timescale initialization. All remaining settings follow
Appendix~\ref{app:train}.

Table~\ref{tab:app-hyp} tests sensitivity to the residual-scale
initialization and graph-base supervision weight. Increasing the
initial residual scale from $\alpha^{R}=0.1$ to $0.3$, or reducing
$\lambda_G$ tenfold from $0.1$ to $0.01$, leaves test nMAE and nRMSE
essentially unchanged across all four protocols.

Table~\ref{tab:app-hyp-rho} reports the same residual-scale and
reliance quantities as Table~\ref{tab:residual} under these
perturbations. A larger initialization produces a substantially
larger converged $\alpha^{R}$, but the realized reliance $\rho_R$
remains nearly unchanged. Reducing
$\lambda_G$ likewise has little effect on either forecasting error or
residual reliance. Across all settings, $\rho_R$ remains in the narrow
range $0.014$--$0.040$.

These results distinguish the learned scale parameter from realized
residual magnitude: different values of $\alpha^{R}$ can produce
similar $\rho_{R}$. The small and stable $\rho_{R}$ shows that the
residual remains small in latent magnitude relative to the graph
predictive base.

\begin{table*}[ht]
  \centering
  \caption{Sensitivity to residual-scale initialization and graph-base
  supervision. \frgr{} uses $\alpha^{R}$ init $0.1$ and
  $\lambda_G=0.1$.}
  \label{tab:app-hyp}
  \small
  \setlength{\tabcolsep}{3.6pt}
  \resizebox{\textwidth}{!}{%
  \begin{tabular}{@{}lrrrrrrrr@{}}
    \toprule
    & \multicolumn{2}{c}{Greenhouse} & \multicolumn{2}{c}{Wastewater}
      & \multicolumn{2}{c}{VitalDB} & \multicolumn{2}{c}{PreDist$^{\dagger}$} \\
    \cmidrule(lr){2-3} \cmidrule(lr){4-5}
    \cmidrule(lr){6-7} \cmidrule(lr){8-9}
    Setting & nMAE & nRMSE & nMAE & nRMSE & nMAE & nRMSE & nMAE & nRMSE \\
    \midrule
    \frgr{}
      & \pmstd{0.1285}{0.0016} & \pmstd{0.1928}{0.0033}
      & \pmstd{0.1245}{0.0018} & \pmstd{0.2157}{0.0028}
      & \pmstd{0.0580}{0.0003} & \pmstd{0.1032}{0.0001}
      & \pmstd{0.0495}{0.0003} & \pmstd{0.0768}{0.0004} \\
    $\alpha^{R}$ init $0.3$
      & \pmstd{0.1296}{0.0014} & \pmstd{0.1930}{0.0017}
      & \pmstd{0.1246}{0.0019} & \pmstd{0.2158}{0.0027}
      & \pmstd{0.0579}{0.0003} & \pmstd{0.1032}{0.0000}
      & \pmstd{0.0502}{0.0009} & \pmstd{0.0773}{0.0009} \\
    $\lambda_G=0.01$
      & \pmstd{0.1272}{0.0004} & \pmstd{0.1915}{0.0009}
      & \pmstd{0.1242}{0.0004} & \pmstd{0.2151}{0.0008}
      & \pmstd{0.0580}{0.0003} & \pmstd{0.1032}{0.0001}
      & \pmstd{0.0497}{0.0009} & \pmstd{0.0768}{0.0010} \\
    \bottomrule
  \end{tabular}}
\end{table*}

\begin{table*}[ht]
  \centering
  \caption{Converged residual scale $\alpha^{R}$ and reliance
  $\rho_R$ (Eq.~\ref{eq:rho}).}
  \label{tab:app-hyp-rho}
  \small
  \setlength{\tabcolsep}{3.6pt}
  \resizebox{\textwidth}{!}{%
  \begin{tabular}{@{}lrrrrrrrr@{}}
    \toprule
    & \multicolumn{2}{c}{Greenhouse} & \multicolumn{2}{c}{Wastewater}
      & \multicolumn{2}{c}{VitalDB} & \multicolumn{2}{c}{PreDist$^{\dagger}$} \\
    \cmidrule(lr){2-3} \cmidrule(lr){4-5}
    \cmidrule(lr){6-7} \cmidrule(lr){8-9}
    Setting & $\alpha^{R}$ & $\rho_R$ & $\alpha^{R}$ & $\rho_R$
      & $\alpha^{R}$ & $\rho_R$ & $\alpha^{R}$ & $\rho_R$ \\
    \midrule
    \frgr{}
      & \pmstd{0.107}{0.001} & \pmstd{0.0302}{0.0019}
      & \pmstd{0.104}{0.001} & \pmstd{0.0391}{0.0032}
      & \pmstd{0.050}{0.002} & \pmstd{0.0142}{0.0009}
      & \pmstd{0.069}{0.001} & \pmstd{0.0143}{0.0016} \\
    $\alpha^{R}$ init $0.3$
      & \pmstd{0.305}{0.002} & \pmstd{0.0274}{0.0030}
      & \pmstd{0.287}{0.002} & \pmstd{0.0383}{0.0012}
      & \pmstd{0.084}{0.005} & \pmstd{0.0141}{0.0010}
      & \pmstd{0.138}{0.002} & \pmstd{0.0156}{0.0007} \\
    $\lambda_G=0.01$
      & \pmstd{0.108}{0.001} & \pmstd{0.0306}{0.0029}
      & \pmstd{0.105}{0.001} & \pmstd{0.0401}{0.0029}
      & \pmstd{0.050}{0.002} & \pmstd{0.0148}{0.0006}
      & \pmstd{0.072}{0.003} & \pmstd{0.0156}{0.0011} \\
    \bottomrule
  \end{tabular}}
\end{table*}

Table~\ref{tab:app-la} tests whether initializing the current
latent-effect belief requires driver history preceding the observation
context. We set $L_A=2L_X$, adding one full context length of past
drivers before the observed history.

Test nMAE and nRMSE remain essentially unchanged on all four
protocols. Additional pre-context driver history therefore provides
no measurable forecasting benefit. The main model consequently uses
context-aligned belief initialization, $L_A=L_X$, rather than a
separate latent-dynamics warm-up window.

\begin{table*}[ht]
  \centering
  \caption{Sensitivity to the latent-effect history window.
  \frgr{} uses context-aligned belief initialization with $L_A=L_X$;
  the ablation adds one additional context length of pre-context driver
  history.}
  \label{tab:app-la}
  \small
  \setlength{\tabcolsep}{3.6pt}
  \resizebox{\textwidth}{!}{%
  \begin{tabular}{@{}lrrrrrrrr@{}}
    \toprule
    & \multicolumn{2}{c}{Greenhouse} & \multicolumn{2}{c}{Wastewater}
      & \multicolumn{2}{c}{VitalDB} & \multicolumn{2}{c}{PreDist$^{\dagger}$} \\
    \cmidrule(lr){2-3} \cmidrule(lr){4-5}
    \cmidrule(lr){6-7} \cmidrule(lr){8-9}
    Setting & nMAE & nRMSE & nMAE & nRMSE & nMAE & nRMSE & nMAE & nRMSE \\
    \midrule
    \frgr{} ($L_A=L_X$)
      & \pmstd{0.1285}{0.0016} & \pmstd{0.1928}{0.0033}
      & \pmstd{0.1245}{0.0018} & \pmstd{0.2157}{0.0028}
      & \pmstd{0.0580}{0.0003} & \pmstd{0.1032}{0.0001}
      & \pmstd{0.0495}{0.0003} & \pmstd{0.0768}{0.0004} \\
    $L_A=2L_X$
      & \pmstd{0.1282}{0.0023} & \pmstd{0.1918}{0.0014}
      & \pmstd{0.1254}{0.0013} & \pmstd{0.2173}{0.0017}
      & \pmstd{0.0580}{0.0003} & \pmstd{0.1032}{0.0001}
      & \pmstd{0.0500}{0.0010} & \pmstd{0.0771}{0.0007} \\
    \bottomrule
  \end{tabular}}
\end{table*}

Table~\ref{tab:app-tau} tests whether the reported forecasting gains
depend on the hand-specified family timescales of
Appendix~\ref{app:train}. We replace all family-specific
initializations with a shared $\tau=8$, corresponding to
$\lambda_f=e^{-1/8}$. Test nMAE and nRMSE remain essentially
unchanged from Table~\ref{tab:main} across all four protocols.

To examine whether training recovers the same persistence parameters
from the two initializations, we convert the learned
$\lambda_f$ back to an implied timescale,
\[
    \hat{\tau}_f = -1/\log \lambda_f ,
\]
the inverse of the initialization
$\lambda_f=e^{-1/\tau_f}$. Table~\ref{tab:app-tau-hat} reports these
values for every latent-effect family.

The two initializations do not in general converge to a common set of
timescales. Several families retain a strong dependence on their
initialization, although some move substantially and others converge
to similar values. Despite these different persistence solutions,
forecast accuracy is nearly unchanged. Thus, on the real forecasting
protocols, the supervised forecasting objective does not uniquely
identify the family persistence parameters: multiple latent dynamics
can support essentially the same predictive solution. The
family-specific $\tau_f$ should therefore be interpreted as an
inductive bias rather than as a recovered physical timescale.

Section~\ref{sec:results-synth} and Appendix~\ref{app:synthetic} test
the same future-rollout mechanism on a matched leaky-integrator
system. Starting from a shared initialization, the model recovers the
correct persistence ordering $\lambda_{\mathrm{slow}}>\lambda_{\mathrm{fast}}$
and exhibits pathway-specific impulse responses. This provides a
controlled capacity check under matched dynamics. On the four real
forecasting protocols, different persistence configurations yield
nearly identical forecast errors, so predictive accuracy alone does
not establish identification of physical timescales.

Accordingly, \method{} does not rely on accurately specifying the
physical family timescales to obtain its reported forecasting
performance: replacing the family-specific initialization by the
shared $\tau=8$ initialization yields essentially the same error.

\begin{table*}[ht]
  \centering
  \caption{Robustness to family-timescale initialization.
  \frgr{} uses the family-specific $\tau_f$ values of
  Appendix~\ref{app:train}; the ablation initializes every family with
  the same $\tau=8$. Forecast accuracy is essentially unchanged.}
  \label{tab:app-tau}
  \small
  \setlength{\tabcolsep}{3.6pt}
  \resizebox{\textwidth}{!}{%
  \begin{tabular}{@{}lrrrrrrrr@{}}
    \toprule
    & \multicolumn{2}{c}{Greenhouse} & \multicolumn{2}{c}{Wastewater}
      & \multicolumn{2}{c}{VitalDB} & \multicolumn{2}{c}{PreDist$^{\dagger}$} \\
    \cmidrule(lr){2-3} \cmidrule(lr){4-5}
    \cmidrule(lr){6-7} \cmidrule(lr){8-9}
    Setting & nMAE & nRMSE & nMAE & nRMSE & nMAE & nRMSE & nMAE & nRMSE \\
    \midrule
    \frgr{}
      & \pmstd{0.1285}{0.0016} & \pmstd{0.1928}{0.0033}
      & \pmstd{0.1245}{0.0018} & \pmstd{0.2157}{0.0028}
      & \pmstd{0.0580}{0.0003} & \pmstd{0.1032}{0.0001}
      & \pmstd{0.0495}{0.0003} & \pmstd{0.0768}{0.0004} \\
    shared $\tau=8$
      & \pmstd{0.1279}{0.0019} & \pmstd{0.1918}{0.0038}
      & \pmstd{0.1244}{0.0011} & \pmstd{0.2150}{0.0021}
      & \pmstd{0.0580}{0.0003} & \pmstd{0.1032}{0.0001}
      & \pmstd{0.0497}{0.0016} & \pmstd{0.0773}{0.0021} \\
    \bottomrule
  \end{tabular}}
\end{table*}

\begin{table}[ht]
  \centering
  \caption{Learned implied family timescales
  $\hat{\tau}_f=-1/\log\lambda_f$ under family-specific and shared
  initialization. The two initializations do not generally converge to
  the same persistence solution despite nearly identical forecasting
  accuracy in Table~\ref{tab:app-tau}.}
  \label{tab:app-tau-hat}
  \footnotesize
  \setlength{\tabcolsep}{4pt}
  \begin{tabular}{@{}llccc@{}}
    \toprule
    Dataset & Family & Prior $\tau_f$
      & $\hat\tau$ (\frgr{})
      & $\hat\tau$ ($\tau=8$) \\
    \midrule
    Greenhouse & thermal & $24$
      & \pmstd{21.4}{0.6} & \pmstd{7.0}{0.1} \\
    & ventilation & $6$
      & \pmstd{5.7}{0.2} & \pmstd{7.4}{0.5} \\
    & light & $2$
      & \pmstd{2.0}{0.0} & \pmstd{8.5}{0.1} \\
    & screen & $4$
      & \pmstd{4.1}{0.1} & \pmstd{8.1}{0.1} \\
    & CO$_2$ & $6$
      & \pmstd{6.3}{0.2} & \pmstd{8.2}{0.3} \\
    \addlinespace
    Wastewater & dosing & $8$
      & \pmstd{9.2}{0.5} & \pmstd{9.0}{1.0} \\
    & aeration & $6$
      & \pmstd{4.3}{0.2} & \pmstd{7.3}{2.8} \\
    & phase & $15$
      & \pmstd{11.3}{0.3} & \pmstd{6.5}{0.3} \\
    \addlinespace
    VitalDB & propofol & $24$
      & \pmstd{19.7}{1.3} & \pmstd{11.8}{0.3} \\
    & remifentanil & $12$
      & \pmstd{10.6}{1.1} & \pmstd{9.1}{1.1} \\
    \addlinespace
    PreDist$^{\dagger}$ & valve & $6$
      & \pmstd{3.8}{0.8} & \pmstd{4.4}{0.5} \\
    & setpoint & $18$
      & \pmstd{6.1}{0.9} & \pmstd{3.8}{0.8} \\
    & pump & $1$
      & \pmstd{1.2}{0.0} & \pmstd{9.1}{0.5} \\
    \bottomrule
  \end{tabular}
\end{table}

Taken together, these ablations show that the reported forecasting
gains are not tied to a narrow optimization recipe. Residual reliance
remains small under different residual initialization and graph-base
supervision, additional pre-context driver history is unnecessary,
and forecast accuracy is robust to replacing family-specific
persistence initialization with a shared timescale.

\section{Synthetic latent-dynamics diagnostics}
\label{app:synthetic}

Section~\ref{sec:results-synth}, Figure~\ref{fig:synth}, and
Table~\ref{tab:synth-impulse} report a class-matched sanity check on
a known instance of Eq.~(\ref{eq:effect-dyn}). This appendix specifies
the leaky-integrator system, the probe protocol, and the full
numerical tables, then examines a saturating $\tanh$ transition mismatch
(Appendix~\ref{app:synthetic-mismatch}).
The probes are the learned family persistences $\lambda_f$ and the
future-impulse map from each driver to each target. Results use
seed $42$.

\subsection{Matched-dynamics system}
\label{app:synthetic-system}

Three continuous drivers $u_{\mathrm{fast}}$, $u_{\mathrm{slow}}$,
and $u_{\mathrm{null}}$ update three latent-effect families. Two
observed targets read a sparse gain:
\begin{equation}
\label{eq:synth-dyn}
  z_{f,t+1}
  =
  \lambda_f\,z_{f,t}
  +
  (1-\lambda_f)\,\alpha_f\,u_{\mathrm{parent}(f),t},
  \qquad
  y_{t}
  =
  W z_{t}
  +
  \varepsilon_{t}.
\end{equation}
Equation~(\ref{eq:synth-dyn}) is the continuous term of
Eq.~(\ref{eq:effect-dyn}) with scalar $(\lambda_f,\alpha_f)$ and
delay $d{=}0$. The ground-truth parameters are
$\lambda_{\mathrm{fast}}=0.40$,
$\lambda_{\mathrm{slow}}=0.75$,
$\alpha_{\mathrm{fast}}=\alpha_{\mathrm{slow}}=1$,
and $\lambda_{\mathrm{null}}=\alpha_{\mathrm{null}}=0$.
$W$ maps the fast effect onto $y_{\mathrm{fast}}$ and the slow
effect onto $y_{\mathrm{slow}}$. The null family has no target edge.
Observation noise is $\varepsilon_t\sim\mathcal{N}(0,0.015^{2})$.
Drivers are a weak AR background with sparse pulses.

The instantaneous step gains are therefore
$(1-\lambda)\alpha=0.60$ on the fast channel and $0.25$ on the slow
channel. 
Both step gains are large relative to the observation-noise scale, so
neither pathway is hidden by noise; the two mechanisms are instead
distinguished primarily by their persistence. 
A unit pulse on $u_{\mathrm{fast}}$ peaks at
$y_{\mathrm{fast}}=0.60$ and leaves $y_{\mathrm{slow}}=0$; a unit
pulse on $u_{\mathrm{slow}}$ peaks at $y_{\mathrm{slow}}=0.25$ and
leaves $y_{\mathrm{fast}}=0$; a pulse on $u_{\mathrm{null}}$ moves
neither target. The typed prior supplied to the model is this
ground-truth graph.

\subsection{Identification protocol}
\label{app:synthetic-protocol}

Windows use context $L_X=8$, horizon $H=32$, and a $64$-step latent
warmup. All but the last context target are zeroed. Superscript
$\mathrm{id}$ marks this identification setting: graph variants use
the hard-bottleneck fusion of
Section~\ref{sec:results-routing} (graph-covered driver tokens
are withheld from the direct path),
family states enter at $\gamma_f{\equiv}1$, and each $\lambda_f$ is
a scalar initialized at $0.575$, the midpoint of $(0.40,0.75)$.
Correct ranking therefore requires the two persistences to move in
opposite directions. \frid{} (\hb{}) is future latent-effect
rollout under that fusion.

The ladder is \dcid{}, future-driver tokens without a graph;
\ieid{}, an instant graph with $\lambda_f{\equiv}0$;
\rbelid{} and \rbelgrid{}, a current recurrent belief without and
with graph-first residual routing; \frid{} (\hb{}) and
\frgrid{}, future latent-effect rollout without and with that
routing. Architecture and optimization follow
Appendix~\ref{app:train} (width $128$, batch $16$, learning rate
$3{\times}10^{-4}$, $40$ epochs). Results are seed $42$. The
training split has $96$ sessions of length $384$.

After training, two probes are read from the checkpoint. The first
is the learned $(\lambda_{\mathrm{fast}},\lambda_{\mathrm{slow}})$
and whether $\lambda_{\mathrm{slow}}>\lambda_{\mathrm{fast}}$. The
second is a unit future pulse on one driver with a zero context: the
reported impulse $I_{d\to y}$ is the maximum absolute response of
target $y$ over the horizon. The ground-truth on-path step gains are $0.60$ and $0.25$; 
off-path entries are the crosstalk and the null
driver.

\subsection{Results}
\label{app:synthetic-results}

Figure~\ref{fig:synth} and Table~\ref{tab:synth-impulse} summarize
learned persistences and future-pulse
responses for the current-belief and rollout variants.
Tables~\ref{tab:app-synth-lambda}--\ref{tab:app-synth-impulse} report
the full identification ladder.

\begin{table}[ht]
  \centering
  \caption{Learned persistence and test MAE on the matched-dynamics
  system (seed $42$). Rank is
  $\lambda_{\mathrm{slow}}>\lambda_{\mathrm{fast}}$.
  \ieid{} fixes $\lambda_f{=}0$. \dcid{} has no family persistence.
  \frid{} (\hb{}) is the hard bottleneck of
  Section~\ref{sec:results-routing}.
  MAE is on the physical target scale.}
  \label{tab:app-synth-lambda}
  \footnotesize
  \setlength{\tabcolsep}{3.6pt}
  \begin{tabular}{@{}lcccccc@{}}
    \toprule
    Method
      & $\lambda_{\mathrm{fast}}$
      & $\lambda_{\mathrm{slow}}$
      & Rank
      & MAE
      & MAE$_{\mathrm{fast}}$
      & MAE$_{\mathrm{slow}}$ \\
    \midrule
    Truth
      & $0.40$ & $0.75$ & ---
      & --- & --- & --- \\
    Init
      & $0.575$ & $0.575$ & ---
      & --- & --- & --- \\
    \midrule
    \dcid{}
      & --- & --- & ---
      & $0.063$ & $0.077$ & $0.049$ \\
    \ieid{}
      & $0$ & $0$ & No
      & $0.063$ & $0.077$ & $0.049$ \\
    \rbelid{}
      & $0.619$ & $0.627$ & No
      & $0.063$ & $0.077$ & $0.049$ \\
    \rbelgrid{}
      & $0.629$ & $0.637$ & No
      & $0.063$ & $0.077$ & $0.049$ \\
    \frid{} (\hb{})
      & $0.449$ & $0.641$ & Yes
      & $0.031$ & $0.013$ & $0.049$ \\
    \frgrid{}
      & $0.429$ & $0.679$ & Yes
      & $0.013$ & $0.013$ & $0.014$ \\
    \bottomrule
  \end{tabular}
\end{table}

\begin{table}[ht]
  \centering
  \caption{Peak absolute response to a unit future pulse
  (horizon $32$, zero context). Truth peaks are $0.60$
  ($u_{\mathrm{fast}}{\to}y_{\mathrm{fast}}$) and $0.25$
  ($u_{\mathrm{slow}}{\to}y_{\mathrm{slow}}$).}
  \label{tab:app-synth-impulse}
  \footnotesize
  \setlength{\tabcolsep}{3.4pt}
  \begin{tabular}{@{}lcccc@{}}
    \toprule
    Method
      & $I_{\mathrm{fast}\to\mathrm{fast}}$
      & $I_{\mathrm{fast}\to\mathrm{slow}}$
      & $I_{\mathrm{slow}\to\mathrm{slow}}$
      & $I_{\mathrm{null}}$ \\
    \midrule
    Truth
      & $0.600$ & $0$ & $0.250$ & $0$ \\
    \dcid{}
      & $0.000$ & $0.000$ & $0.000$ & $0.000$ \\
    \ieid{}
      & $0.000$ & $0.000$ & $0.000$ & $0.000$ \\
    \rbelid{}
      & $0.000$ & $0.000$ & $0.000$ & $0.000$ \\
    \rbelgrid{}
      & $0.000$ & $0.000$ & $0.000$ & $0.000$ \\
    \frid{} (\hb{})
      & $0.371$ & $0.010$ & $0.000$ & $0.001$ \\
    \frgrid{}
      & $0.430$ & $0.008$ & $0.141$ & $0.002$ \\
    \bottomrule
  \end{tabular}
\end{table}

The no-rollout variants fail to recover horizon-dependent responses
to future driver pulses. \dcid{}, \ieid{}, \rbelid{}, and
\rbelgrid{} all record MAE $0.063$, with
$y_{\mathrm{fast}}$ MAE $0.077$ and $y_{\mathrm{slow}}$ MAE
$0.049$, and all four produce zero future-pulse responses.
\ieid{} fixes $\lambda_f{\equiv}0$, and the two current-belief
models leave the persistences nearly equal
($\lambda{\approx}0.62$--$0.64$) rather than separating the fast
and slow families. \dcid{} receives future-driver tokens directly,
but likewise produces zero pulse responses. Thus, under this
identification protocol, direct future conditioning alone does not
recover the latent future driver-response dynamics.

The current graph belief $H_t^{G}$ is computed at the forecast
anchor and remains fixed over the prediction horizon. A driver
pulse introduced at lead time $h$ therefore never updates $Z^{F}$
in the current-belief variants, so $\lambda_f$ receives no
gradient that identifies how later driver effects should persist
across future horizons. This is why \rbelid{} and \rbelgrid{}
match on every probe.

Future rollout changes this identification path. Applying
$Z_{t+h}^{F}=\mathcal{F}(Z_{t+h-1}^{F},D_{t+h-1})$ at every lead time causes a later
driver pulse to enter $Z_{t+h}^{F}$ and propagate across
subsequent horizons under $\lambda_f$. \frid{} (\hb{}) moves
$\lambda_{\mathrm{fast}}$ from $0.575$ to $0.449$ and
$\lambda_{\mathrm{slow}}$ from $0.575$ to $0.641$, recovering
the correct persistence ordering and reducing MAE from $0.063$
to $0.031$. Its fast-path pulse response rises to $0.371$ versus
the ground-truth $0.60$, with only $0.010$ crosstalk onto
$y_{\mathrm{slow}}$. That gain is confined to the fast target:
MAE falls from $0.077$ to $0.013$, while the slow-path pulse
stays at $0$ and $y_{\mathrm{slow}}$ MAE remains $0.049$.

Graph-first residual routing recovers the remaining pathway.
\frgrid{} moves the learned persistences to $0.429$ and $0.679$,
closer to the ground-truth $(0.40,0.75)$, reduces overall MAE to
$0.013$, and lowers the two per-target MAEs to $0.013$ and
$0.014$. The corresponding future-pulse responses are $0.430$ on
the fast wired path and $0.141$ on the slow wired path, while
off-path crosstalk remains $0.008$ and the null-driver response
$0.002$. 
Thus, under matched dynamics, the rollout variants recover the 
correct persistence ordering, while the full \frgrid{} model 
exhibits pathway-specific responses on both wired paths.

\subsection{Saturating \texorpdfstring{$\tanh$}{tanh} mismatch}
\label{app:synthetic-mismatch}

The matched system in Eq.~(\ref{eq:synth-dyn}) uses the same linear
leaky integrator as the model. To test a mild class mismatch we keep
$(\lambda_f,\alpha_f)$, $d{\equiv}0$, observation noise, drivers, and
the typed prior fixed, and replace the latent update by
$z\leftarrow\tanh\bigl(\lambda z+(1-\lambda)\alpha u\bigr)$.
A unit pulse then peaks at $\tanh(0.60)\approx0.537$ on
$y_{\mathrm{fast}}$ and $\tanh(0.25)\approx0.245$ on
$y_{\mathrm{slow}}$.
The trained model remains the linear integrator of
Eq.~(\ref{eq:effect-dyn}).
Optimization and initialization follow
Appendix~\ref{app:synthetic-protocol}
(width $128$, batch $16$, learning rate $3{\times}10^{-4}$,
seed $42$, $\alpha$ initialized at $0.5$).
The identification ladder is trained for $40$ epochs, matching the
linear system; the last row trains the same \frgrid{} recipe for
$80$ epochs.

Tables~\ref{tab:app-synth-nl-lambda}--\ref{tab:app-synth-nl-impulse}
report the same probes as the matched system.

\begin{table}[ht]
  \centering
  \caption{Learned persistence and test MAE on the saturating
  $\tanh$ system (seed $42$). Rank is
  $\lambda_{\mathrm{slow}}>\lambda_{\mathrm{fast}}$.
  \ieid{} fixes $\lambda_f{=}0$. \dcid{} has no family persistence.
  The ladder uses $40$ epochs; the last row is \frgrid{} at
  $80$ epochs. MAE is on the physical target scale.}
  \label{tab:app-synth-nl-lambda}
  \footnotesize
  \setlength{\tabcolsep}{3.6pt}
  \begin{tabular}{@{}lcccccc@{}}
    \toprule
    Method
      & $\lambda_{\mathrm{fast}}$
      & $\lambda_{\mathrm{slow}}$
      & Rank
      & MAE
      & MAE$_{\mathrm{fast}}$
      & MAE$_{\mathrm{slow}}$ \\
    \midrule
    Truth
      & $0.40$ & $0.75$ & ---
      & --- & --- & --- \\
    Init
      & $0.575$ & $0.575$ & ---
      & --- & --- & --- \\
    \midrule
    \dcid{}
      & --- & --- & ---
      & $0.044$ & $0.056$ & $0.034$ \\
    \ieid{}
      & $0$ & $0$ & No
      & $0.044$ & $0.055$ & $0.034$ \\
    \rbelid{}
      & $0.622$ & $0.628$ & No
      & $0.044$ & $0.055$ & $0.034$ \\
    \rbelgrid{}
      & $0.632$ & $0.639$ & No
      & $0.044$ & $0.055$ & $0.034$ \\
    \frid{} (\hb{})
      & $0.580$ & $0.591$ & No
      & $0.044$ & $0.055$ & $0.034$ \\
    \frgrid{} ($40$ ep)
      & $0.401$ & $0.609$ & Yes
      & $0.023$ & $0.013$ & $0.034$ \\
    \frgrid{} ($80$ ep)
      & $0.343$ & $0.615$ & Yes
      & $0.012$ & $0.012$ & $0.012$ \\
    \bottomrule
  \end{tabular}
\end{table}

\begin{table}[ht]
  \centering
  \caption{Peak absolute response to a unit future pulse on the
  saturating $\tanh$ system
  (horizon $32$, zero context).
  Truth peaks are $0.537$
  ($u_{\mathrm{fast}}{\to}y_{\mathrm{fast}}$) and $0.245$
  ($u_{\mathrm{slow}}{\to}y_{\mathrm{slow}}$).
  The ladder uses $40$ epochs; the last row is \frgrid{} at
  $80$ epochs.}
  \label{tab:app-synth-nl-impulse}
  \footnotesize
  \setlength{\tabcolsep}{3.4pt}
  \begin{tabular}{@{}lcccc@{}}
    \toprule
    Method
      & $I_{\mathrm{fast}\to\mathrm{fast}}$
      & $I_{\mathrm{fast}\to\mathrm{slow}}$
      & $I_{\mathrm{slow}\to\mathrm{slow}}$
      & $I_{\mathrm{null}}$ \\
    \midrule
    Truth
      & $0.537$ & $0$ & $0.245$ & $0$ \\
    \dcid{}
      & $0.000$ & $0.000$ & $0.000$ & $0.001$ \\
    \ieid{}
      & $0.000$ & $0.000$ & $0.000$ & $0.000$ \\
    \rbelid{}
      & $0.000$ & $0.000$ & $0.000$ & $0.000$ \\
    \rbelgrid{}
      & $0.000$ & $0.000$ & $0.000$ & $0.000$ \\
    \frid{} (\hb{})
      & $0.000$ & $0.000$ & $0.000$ & $0.000$ \\
    \frgrid{} ($40$ ep)
      & $0.278$ & $0.008$ & $0.001$ & $0.000$ \\
    \frgrid{} ($80$ ep)
      & $0.330$ & $0.015$ & $0.126$ & $0.000$ \\
    \bottomrule
  \end{tabular}
\end{table}

Without future rollout the $40$-epoch probes match the linear
system: \dcid{}, \ieid{}, \rbelid{}, and \rbelgrid{} leave MAE at
$0.044$ and produce no future-pulse response, and the
current-belief persistences stay near $0.62$--$0.64$.
Unlike the matched system, \frid{} (\hb{}) also stays near the
initialization ($0.580$, $0.591$), does not recover the rank, and
records zero pulse responses.
At $40$ epochs, \frgrid{} separates the persistences
($0.401<0.609$) and halves MAE to $0.023$ by fitting the fast
target ($0.013$), but $y_{\mathrm{slow}}$ MAE remains $0.034$ and
the slow-path pulse is $0.001$ against the saturated peak $0.245$.
The fast-path pulse is $0.278$ against $0.537$, with $0.008$
crosstalk.

At $80$ epochs the same \frgrid{} recipe recovers both wired
pathways: the slow-path pulse rises to $0.126$, the fast-path
pulse to $0.330$, and the two per-target MAEs both fall to
$0.012$.
The learned persistences are then $0.343$ and $0.615$, still
farther from $(0.40,0.75)$ than the $40$-epoch matched estimates
($0.429$, $0.679$).

\section{Typed-graph generation prompt}
\label{app:graph-prompt}

Cursor Grok~4.6~\citep{cursor2026_grok46} is queried once per dataset,
offline. PreDist receives a second query that tags each edge
$\mathrm{dyn}$ or $\mathrm{alg}$. The input is the schema, variable
descriptions, and the constraints below. The output is a JSON graph that is checked against a
fixed schema, reviewed, and frozen; training loads that file and never
calls the model. Self-loops and reverse message-passing edges are added
programmatically after the proposal. Listing~\ref{lst:graph-prompt-shared}
is the shared instruction. Dataset-specific node inventories and routing
constraints follow. CGMacros and Shanghai Diabetes use the same shared
instruction with a single glucose target; their frozen prompts are stored
with the graphs.

\begin{lstlisting}[style=prompt,caption={Shared prompt for typed-graph generation.},label=lst:graph-prompt-shared]
You are proposing a minimal symbolic relational graph prior for an
action-conditioned world model.

Given the dataset schema and the hard constraints below, return a
machine-readable JSON graph with:
1) nodes: id, display_name, type in {state, action, exo, latent-effect}
2) directed edges: source, target, relation, sign in {positive,
   negative, unknown}, confidence in [0,1], short rationale

Shared constraints:
- Driver nodes (action or exo) update latent-effect nodes; latent-effect
  nodes reach observed targets.
- Do not add direct action->state edges that skip latent-effect nodes.
- Do not add intra-state edges unless a dataset constraint explicitly
  allows an algebraic one.
- All exogenous channels remain available as direct future-driver
  tokens; only the listed subset become graph exo nodes.
- Exo nodes are source-only: do not add reverse message-passing edges
  into exo.
- Use the shared relation 'activates' for action-driver and exo-driver
  edges into effect nodes unless a dataset constraint assigns a unique
  relation id.
- Give each sign=unknown action-modulates edge a unique relation id so
  the GNN does not share W_r across them.
- Self-loops and reverse message-passing edges are added
  programmatically later; do not invent extra causal claims.
- Output JSON only, matching the repository schema.
\end{lstlisting}

\textbf{VitalDB.}
Intraoperative vitals on a regularly sampled grid: MBP, HR, SpO$_2$, BIS.
Dense continuous actions: propofol and remifentanil pump rates. No
exogenous variables.

\begin{lstlisting}[style=prompt,caption={VitalDB graph constraints.}]
Hard constraints:
- Nodes: MBP, HR, SpO2, BIS (state); Propofol, Remifentanil (action);
  Propofol Effect, Remifentanil Effect (latent-effect).
- Edges: each drug activates its own effect; Propofol Effect decreases
  BIS, MBP, HR; Remifentanil Effect decreases BIS, MBP, HR.
- Do not add a drug-effect edge into SpO2.
- Do not merge the two drugs into one shared effect node.
- Do not create nodes for rate, Delta-rate, or on/off.
\end{lstlisting}

\textbf{Greenhouse.}
Indoor climate on a 5\,min grid: Tair, Rhair, CO$_2$air, PARin. Dense
continuous actions: Tpipe, VentLee, VentWind, AssimLight, EnScr,
BlckScr, CO$_2$dosing. All eight outdoor-weather channels stay on the
direct path; only Tout, Windsp, PARout, and AbsHumOut are graph exo
nodes.

\begin{lstlisting}[style=prompt,caption={Greenhouse graph constraints.}]
Hard constraints:
- State: tair, rhair, co2air, parin.
- Action drivers: tpipe, vent_lee, vent_wind, assim_light, en_scr,
  blck_scr, co2_dosing.
- Exo drivers: tout, windsp, parout, abshumout (refreshed each step).
- Effects: thermal, ventilation, light, screen, co2.
- Routes: tpipe+Tout -> thermal; vent_lee+vent_wind+Windsp+Tout+AbsHumOut
  -> ventilation; assim_light+PARout -> light;
  en_scr+blck_scr+PARout+Tout -> screen; co2_dosing -> co2.
- Do not add Rhout, Iglob, Winddir, or Rain as graph nodes.
- Do not add dense exo-to-target edges.
\end{lstlisting}

\textbf{Wastewater nutrient control.}
Tank-1 nutrient state on a 2\,min grid: DO, NH$_4$, PO$_4$. Continuous
actions: metal-salt dosing and blower factor. Categorical actions: inlet
and outlet process phases. Exogenous: inflow and water temperature; both
are graph exo nodes.

\begin{lstlisting}[style=prompt,caption={Wastewater graph constraints.}]
Hard constraints:
- State: do_tank1, nh4, po4.
- Action drivers: inlet_metal_dosing, metal_dosing, max_chem_factor,
  processphase_inlet, processphase_outlet.
- Exo drivers: inflow, temperature (refreshed each step).
- Effects: dosing, aeration, phase.
- Routes: inlet_metal_dosing+metal_dosing+temperature -> dosing;
  max_chem_factor+inflow+temperature -> aeration;
  processphase_inlet+processphase_outlet+inflow -> phase.
- Do not add a temperature driver into phase_effect.
- Do not treat inflow or temperature as actions.
- Do not add dense exo-to-target edges.
\end{lstlisting}

\textbf{PreDist.}
Substation thermodynamics on a 10\,min grid. Continuous actions: valve
position and supply-temperature setpoint. Categorical actions: pump
status, control mode, three-way valve. Exogenous: outdoor temperature
and primary-grid supply temperature. Temperatures are inertial
($\mathrm{dyn}$); heat power and flow are algebraic ($\mathrm{alg}$).
A second query, given the same nodes, tags every edge with timescale
$\mathrm{dyn}$ or $\mathrm{alg}$ and uses a disjoint relation
vocabulary.

\begin{lstlisting}[style=prompt,caption={PreDist dual-timescale graph constraints.}]
Hard constraints:
- State: s_hc1_supply_temp, p_hc1_return_temp, p_net_return_temp
  (dyn); heat_power, flow (alg).
- Action drivers: valve_pos, supply_setpoint, pump_status,
  control_mode, dhw_3way.
- Exo drivers: outdoor_temp, p_net_supply (refreshed each step).
- Effects: valve, setpoint, pump. Effects drive temperatures only.
- Tag every edge timescale=dyn or timescale=alg. Use disjoint
  relation ids so G_dyn and G_alg do not share W_r.
- G_dyn routes: valve_pos+dhw_3way+p_net_supply+outdoor_temp -> valve;
  supply_setpoint+control_mode+outdoor_temp -> setpoint;
  pump_status -> pump;
  valve -> supply and both returns;
  setpoint -> supply and both returns;
  pump -> supply.
- G_alg routes: valve_pos+dhw_3way+pump_status -> flow (direct
  action->state, skipping latent-effect nodes);
  flow -> heat_power (determines);
  p_net_supply and p_hc1_return_temp -> heat_power (primary dT fan-in).
- Do not add effect -> flow or effect -> heat_power.
- Do not add action -> heat_power.
- Do not add supply_setpoint or control_mode into flow.
- Do not add reverse message-passing on G_alg.
- Do not add a weather driver into pump_effect.
- Do not encode maintenance or faults as actions or graph nodes.
- Do not add dense exo-to-target edges.
- Self-loops and reverse MP (dyn only, never into exo) are added
  programmatically later.
\end{lstlisting}

\end{document}